\documentclass{article}
\usepackage{epsf}
\usepackage{dsfont,natbib}
\usepackage{amsfonts,dsfont,amssymb,amsmath,graphics,graphicx}
\usepackage{color}
\usepackage{epsfig}

\usepackage{dsfont}

\newtheorem{definition}{Definition}

\newtheorem{example}{Example}

\usepackage{enumitem}
\usepackage{enumerate}
\usepackage{amsmath}

\def\RR{\mathbb{R}}

\def\NN{\mathds{N}}

\def\Ecal{\mathcal{E}}

\def\Dcal{\mathcal{D}}
\def\Scal{\mathcal{S}}
\def\Xcal{\mathcal{X}}

\def\Vcal{\mathcal{V}}
\def\Fcal{\mathcal{F}}

\def\Pcal{\mathcal{P}}

\def\Ncal{\mathcal{N}}
\def\Hcal{\mathcal{H}}

\def\Hcalk{\Hcal_k}
\def\Lcal{\mathcal{L}}

\def\Bcal{\mathcal{B}}
\def\Ycal{\mathcal{Y}}
\def\Tcal{\mathcal{T}}

\def\OMIT#1{}

\DeclareMathOperator{\spa}{span} 
 
\DeclareMathOperator{\corr}{corr}
\DeclareMathOperator{\cov}{cov}
\DeclareMathOperator{\defi}{def}

 \DeclareMathOperator{\emp}{emp}
\DeclareMathOperator{\defeq}{\overset{\defi}{=}}

\DeclareMathOperator{\var}{var}
\DeclareMathOperator{\argm}{argmax}

\usepackage{enumerate,multirow}

\usepackage{tabularx}

\makeatletter
\newif\if@borderstar
\def\bordermatrix{\@ifnextchar*{%
  \@borderstartrue\@bordermatrix@i}{\@borderstarfalse\@bordermatrix@i*}%
}
\def\@bordermatrix@i*{\@ifnextchar[{%
  \@bordermatrix@ii}{\@bordermatrix@ii[()]}
}
\def\@bordermatrix@ii[#1]#2{%
  \begingroup
    \m@th\@tempdima8.75\p@\setbox\z@\vbox{%
      \def\cr{\crcr\noalign{\kern 2\p@\global\let\cr\endline }}%
      \ialign {$##$\hfil\kern 2\p@\kern\@tempdima & \thinspace %
      \hfil $##$\hfil && \quad\hfil $##$\hfil\crcr\omit\strut %
      \hfil\crcr\noalign{\kern -\baselineskip}#2\crcr\omit %
      \strut\cr}}%
    \setbox\tw@\vbox{\unvcopy\z@\global\setbox\@ne\lastbox}%
    \setbox\tw@\hbox{\unhbox\@ne\unskip\global\setbox\@ne\lastbox}%
    \setbox\tw@\hbox{%
      $\kern\wd\@ne\kern -\@tempdima\left\@firstoftwo#1%
        \if@borderstar\kern2pt\else\kern -\wd\@ne\fi%
      \global\setbox\@ne\vbox{\box\@ne\if@borderstar\else\kern 2\p@\fi}%
      \vcenter{\if@borderstar\else\kern -\ht\@ne\fi%
        \unvbox\z@\kern-\if@borderstar2\fi\baselineskip}%
        \if@borderstar\kern-2\@tempdima\kern2\p@\else\,\fi\right\@secondoftwo#1 $%
    }\null \;\vbox{\kern\ht\@ne\box\tw@}%
  \endgroup
}
\makeatother

\newcommand{\BEAS}{\begin{eqnarray*}}
\newcommand{\EEAS}{\end{eqnarray*}}
\newcommand{\BEA}{\begin{eqnarray}}
\newcommand{\EEA}{\end{eqnarray}}
\newcommand{\BEQ}{\begin{equation}}
\newcommand{\EEQ}{\end{equation}}
\newcommand{\BIT}{\begin{itemize}}
\newcommand{\EIT}{\end{itemize}}
\newcommand{\BNUM}{\begin{enumerate}}
\newcommand{\ENUM}{\end{enumerate}}

\newcommand{\BA}{\begin{array}}
\newcommand{\EA}{\end{array}}
\newcommand{\BC}{\begin{center}}
\newcommand{\EC}{\end{center}}

\newcommand{\ones}{\mathbf 1}

\newcommand{\diag}{\mathop{\bf diag}}

\newcommand{\argmin}{\mathop{\rm argmin}}

\newtheorem{theorem}{Theorem}

\newtheorem{proposition}[theorem]{Proposition}
\newtheorem{remark}[theorem]{Remark}

\newcounter{exno}

{\begin{quote}}{\end{quote}}

\makeatletter
\long\def\@makecaption#1#2{
   \vskip 9pt
   \begin{small}
   \setbox\@tempboxa\hbox{{\bf #1:} #2}
   \ifdim \wd\@tempboxa > 5.5in
        \begin{center}
        \begin{minipage}[t]{5.5in}
        \addtolength{\baselineskip}{-0.95pt}
        {\bf #1:} #2 \par
        \addtolength{\baselineskip}{0.95pt}
        \end{minipage}
        \end{center}
   \else
    \hbox to\hsize{\hfil\box\@tempboxa\hfil}
   \fi
   \end{small}\par
}
\makeatother

\newcounter{oursection}

\newcounter{lecture}

\title{Positive Definite Kernels in Machine Learning}
\author{Marco Cuturi\footnote{\texttt{mcuturi@princeton.edu}. Currently with ORFE- Princeton University. A large share of this work was carried out while the author was working at the Institute of Statistical Mathematics, Tokyo, Japan. In particular, this research was supported by the Function and Induction Research Project, Transdisciplinary Research Integration Center, Research Organization of Information and Systems.}}
\begin{document}
\maketitle

\begin{abstract}
This survey is an introduction to positive definite kernels and the set of methods they have inspired in the machine learning literature, namely kernel methods. We first discuss some properties of positive definite kernels as well as reproducing kernel Hibert spaces, the natural extension of the set of functions $\{k(x,\cdot),x\in\Xcal\}$ associated with a kernel $k$ defined on a space $\Xcal$. We discuss at length the construction of kernel functions that take advantage of well-known statistical models. We provide an overview of numerous data-analysis methods which take advantage of reproducing kernel Hilbert spaces and discuss the idea of combining several kernels to improve the performance on certain tasks. We also provide a short cookbook of different kernels which are particularly useful for certain data-types such as images, graphs or speech segments.
\end{abstract}

\textbf{{Remark:}} This report is a draft. Comments and suggestions will be highly appreciated.

\section*{Summary}
We provide in this survey a short introduction to positive definite kernels and the set of methods they have inspired in machine learning, also known as kernel methods. The main idea behind kernel methods is the following. Most data-inference tasks aim at defining an appropriate decision function $f$ on a set of objects of interest $\Xcal$. When $\Xcal$ is a vector space of dimension $d$, say $\RR^d$, linear functions $f_a(x)=a^T x$ are one of the easiest and better understood choices, notably for regression, classification or dimensionality reduction. Given a positive definite kernel $k$ on $\Xcal$, that is a real-valued function on $\Xcal\times\Xcal$ which quantifies effectively how similar two points $x$ and $y$ are through the value $k(x,y)$, kernel methods are algorithms which estimate functions $f$ of the form 
\begin{equation}\label{eq:f}
f:x\in\Xcal \rightarrow f(x)=\sum_{i\in I}\alpha_i k(x_i,x),
\end{equation} 
where $(x_i)_{i\in I}$ is a family of known points paired with $(\alpha_i)_{i\in I}$, a family of real coefficients. Kernel methods are often referred to as
\begin{itemize}
\item \emph{data-driven} since the function $f$ described in Equation~\eqref{eq:f} is an expansion of evaluations of the kernel $k$ on points observed in the sample $I$, as opposed to a linear function $a^T x$ which only has $d$ parameters;
\item \emph{non-parametric} since the vector of parameters $(\alpha_i)$ is indexed on a set $I$ which is of variable size;
\item \emph{non-linear} since $k$ can be a non-linear function such as the gaussian kernel $k(x,y)=\exp(-\|x-y\|^2/ (2\sigma^2))$, and result in non-linear compounded functions $f$.
\item \emph{easily handled through convex programming} since many of the optimization problems formulated to propose suitable choices for the weights $\alpha$ involve quadratic constraints and objectives, which typically involve terms of the sort $\alpha^T K \alpha$ where $K$ is a positive semi-definite matrix of kernel evaluations $[k(x_i,x_j)]$.
\end{itemize}

The problem of defining all of the elements introduced above, from the kernel $k$ to the index set $I$ and most importantly the weights $\alpha_i$ has spurred a large corpus of literature. We propose a survey of such techniques in this document. Our aim is to provide both theoretical and practical insights on positive definite kernels.

This survey is structured as follows:
\begin{itemize}
\item We start this survey by giving an overall introduction to kernel methods in Section~\ref{chap:intro} and highlight their specificity.
\item  We provide the reader with the theoretical foundations that underlie positive definite kernels in Section~\ref{chap:mathkernel}, introduce reproducing kernel Hilbert spaces theory and provide a discussion on the relationships between positive definite kernels and distances.
\item Section~\ref{chap:creatingkernels} describes different families of kernels which have been covered in the literature of the last decade. We also describe a few popular techniques to encode prior knowledge on objects when defining kernels. 
\item We follow with the exposition in Section~\ref{chap:machines} of popular methods which, paired with the definition of a kernel, provide estimation algorithms to define the weights $\alpha_i$ of Equation~\eqref{eq:f}. 
\item Selecting the right kernel for a given application is a practical hurdle when applying kernel methods in practice. We provide a few techniques to do so in Section~\ref{chap:interactions}, notably parameter tuning and the construction of linear mixtures of kernels, also known as multiple kernel learning. 
\item We close the survey by providing a brief cookbook of kernels in Section~\ref{chap:cookbook}, that is a short description of kernels for complex objects such as strings, texts, graphs and images.
\end{itemize}
This survey is built on earlier references, notably \citep{schoelkopf02learning,schoelkopf04kernel,shawe2004kernel}. Whenever adequate we have tried to enrich this presentation with slightly more theoretical insights from~\citep{berg84harmonic,berlinet03reproducing}, notably in Section~\ref{chap:mathkernel}. Topics covered in this survey overlap with some of the sections of~\citep{muller2001introduction} and more recently~\citep{hofmann2008kernel}. The latter references cover in more detail kernel machines, such as the support vector machine for binary or multi-class classification. This presentation is comparatively tilted towards the study of positive definite kernels, notably in Sections~\ref{chap:mathkernel} and ~\ref{chap:creatingkernels}
\newpage
\tableofcontents
\newpage
\section{Introduction}\label{chap:intro}
The automation of data collection in most human activities, from industries, public institutions to academia, has generated tremendous amounts of observational data. In the same time, computational means have expanded in such a way that massive
parallel clusters are now an affordable commodity for most laboratories and small companies. Unfortunately, recent years have seen an increasing gap of efficiency between our ability to produce and store these databases and our the analytical tools that are needed to infer knowledge from them. This long quest to understand and analyze such databases has spurred in the last decades fertile discoveries at the intersection of mathematics, statistics and computer science.

One of the most interesting changes brought forward by the abundance of data in recent years lies arguably in the increasing diversity of data structures practitioners are now faced with. Some complex data types that come
from real-life applications do not translate well into simple vectors of features, which used to be a \emph{de facto} requirement for statistical analysis up to four decades ago. When the task on such data types can be translated into elementary
subtasks that involve for instance \emph{regression, binary or multi-class
classification, dimensionality reduction, canonical correlation
analysis} or \emph{clustering}, a novel class of algorithms popularized in the late nineties and known as
kernel methods have proven to be effective, if not reach
state-of-the art performance on many of these problems.

\paragraph{statistics, functional analysis and computer science:}
the mathematical machinery of kernel methods can be traced back to
the seminal presentation of reproducing kernel Hilbert spaces
by~\citet{aron50} and its use in non-parametric statistics
by~\citet{Parzen:1962:EDP}. However, their recent popularity in machine learning comes from recent innovations in both the \emph{design of kernels} geared towards specific applications such as the one we cover in Section~\ref{chap:cookbook}, paired with efficient \emph{kernel machines} as introduced in Section~\ref{chap:machines}. Examples of the latter include algorithms such as gaussian processes with sparse
representations~\citep{Csato+Opper:2002} or the popular support
vector machine~\citep{mach:Cortes+Vapnik:1995}. The theoretical justifications for such tools can be found in the statistical learning literature~\citep{CucSma02,vapnik98statistical} but also in subsequent convergence and consistency analysis carried out for specific techniques~\citep{fukumizu-kcca,vertvert,bach2008consistency}.
Kernel design embodies the research trend pionneered in~\citet{jaakkola-haussler-99,haussler99convolution,watkins00dynamic} of  incorporating contextual knowledge on the objects of interest to define kernels.

Two features of kernel methods have been often quoted to explain the practical success of kernel methods. First, kernel methods can handle efficiently complex data types through the definition of appropriate kernels. Second, kernel methods can handle data which have multiple data representations, namely multimodal data. Let us review these claims before introducing the mathematical definition of kernels in the next section.

\subsection{Versatile Framework for Structured Data}
Structured objects such as (to cite a few) strings, 3D structures,
trees and networks, time-series, histograms, images, and texts
have become in an increasing number of applications the \textit{de
facto} inputs for data analysis algorithms. The originality of kernel methods is to address this diversity through a single approach.

\begin{figure}\label{fig:mapping}
\begin{center}
\scalebox{.7}{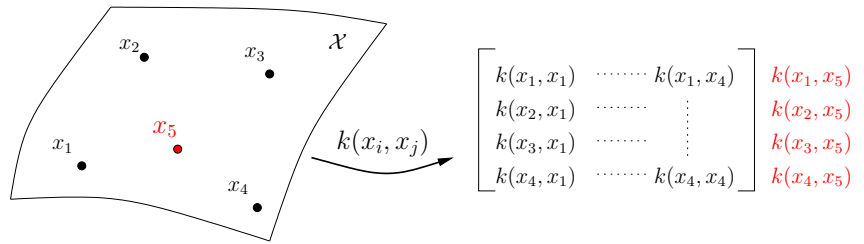}
\caption{Given a dataset in a given space $\Xcal$, represented as $\{x_1,x_2,x_3,x_4\}$ in the figure above, the kernel approach to data analysis involves representing these points through a positive-definite symmetric matrix of inter-similarities between points, as in the matrix $K_{4\times 4}$ in the figure on the right. Given a new point $x_5$, any prediction with respect to $x_5$ (as in regression or classification for instance) will be a direct function of the similarity of $x_5$ to the learning set $\{x_1,x_2,x_3,x_4\}$. Thus, and in practice, kernel methods rely exclusively, both in the training phase and the actual use of the decision function, on similarity matrices.}
\end{center}
\end{figure}

\paragraph{from $n$ points to $n\times n$ similarity matrices:}
using kernel methods on a dataset usually involves choosing first a family of similarity
measures between pairs of objects. Irrespective of the initial complexity of the considered
objects, dealing with a learning problem through
kernels is equivalent to translating a set of $n$ data points into a
symmetric and positive definite $n\times n$ similarity matrix. This matrix will be the sole input used by the kernel algorithm, as schematically shown on Figure~\ref{fig:mapping}. This is very similar to the k-nearest neighbor (k-NN) framework (see~[\S13]\citep{hastie01} for a survey) where only distances between points matter to derive decision functions. On the contrary, parametric approaches used in statistics and neural networks impose a functional class beforehand (e.g. a family of statistical models or a neural
architecture), which is either tailored to fit vectorial data --
which in most cases requires a feature extraction procedure to avoid large
or noisy vectorial representations -- or tailored to fit a
particular data type (hidden Markov models with strings, Markov
random fields with images, parametric models for time series with given lags and seasonal corrections etc.). In this context, practitioners usually give kernel methods different credits, among them the fact that
\begin{itemize}
\item Defining kernel functions is in general easier that designing an accurate generative model and the estimation machinery that goes along with it, notably the optimization mechanisms and/or bayesian computational schemes that are necessary to make computations tractable.
\item Efficient kernel machines, that is algorithm which use directly as an input Kernel matrices, such as the SVM or kernel-PCA, are numerous and the subject of separate research. Their wide availability under the form of software packages, makes them simple to use once a kernel has been defined.
\item Kernel methods share initially the conceptual simplicity of k-nearest neighbors  which make them popular when dealing with high-dimensional and challenging datasets for which little is known beforehand, such as the study of long sequences in bioinformatics~\cite{vert2006classification}. On the other hand, kernel algorithms offer a wider scope than the regression/classification applications of k-NN and also provide a motivated answer to control the bias/variance tradeoff of the decision function through penalized estimation, as explained in Section~\ref{sec:supervized}.
\end{itemize}

\subsection{Multimodality and Mixtures of Kernels}
In most applications currently studied by practitioners, datasets are increasingly \emph{multimodal}. Namely, described objects of interest through the lens of different representations.

\begin{figure}\label{fig:multiplekernel}
\begin{center}
\scalebox{.8}{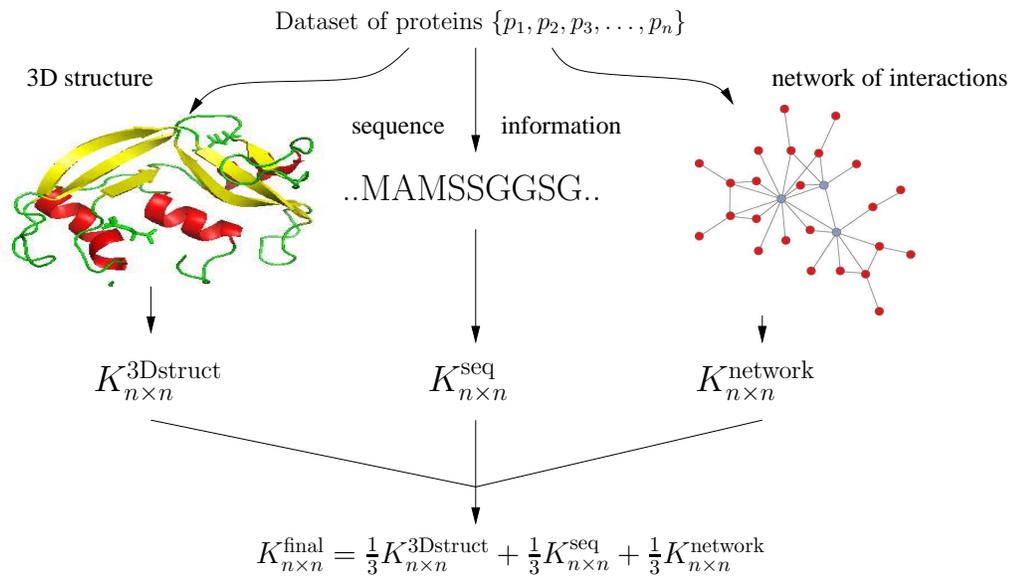}
\caption{A dataset of proteins can be regarded in (at least) three different ways: as a dataset of 3D structures, a dataset of sequences and a set of nodes in a network which interact with each other. A different kernel matrix can be extracted from each datatype, using known kernels on 3D shapes, strings and graphs. The resulting kernels can then be combined together with arbitrary weights, as is the case above where a simple average is considered, or estimated weights, which is the subject of Section~\ref{subsec:addmix}}
\end{center}
\end{figure}

For instance, a protein can be seen as an amino-acid sequence, a macro-molecule with a 3D-structure,
an expression level in a DNA-chip, a node in a biological pathway
or in a phylogenetic tree. A video segment might be characterized by its images, its soundtrack, or additional information such as when it was broadcasted and on which channel. The interrelations between these
modalities and the capacity to integrate them is likely to prove helpful for most learning tasks. Kernel methods provide an elegant way of integrating multimodalities through convex kernel
combinations. This combination takes usually place before using a kernel machine as illustrated in Figure~\ref{fig:multiplekernel}. This stands in stark contrast to other standard techniques which usually aggregate decision functions trained on the separated modalities. A wide range of techniques have been designed to do so through convex
optimization and the use of unlabelled data~\citep{lanckriet2004,Sindhwani}. Kernels can thus be seen as atomic elements that focus on certain types of similarities for the objects, which can be combined through so-called multiple kernel learning methods as will be exposed more specifically in Section~\ref{subsec:addmix}.

\newpage\section{Kernels: a Mathematical Definition}\label{chap:mathkernel}
\subsection{Positive Definiteness}
Let us start this section by providing the reader with a definition
for kernels, since the term ``kernel'' itself is used in different
branches of mathematics, from linear algebra, density estimation to
integral operators theory. Some classical kernels used in non-parametric statistics, such as the Epanechnikov kernel\footnote{for $h>0$,
$k_h(x,y)=\frac{3}{4}\left(1-\left(\frac{x-y}{h}\right)^2\right)^+$},
are not, for instance, kernels in the sense of the terminology
adopted in this report. We develop in this section elementary insights
on kernels, combining different presentations given in
~\citep{berlinet03reproducing,berg84harmonic,schoelkopf02learning}
to which the reader may refer for a more complete exposition.

\paragraph{basic mathematic definition:}
let $\Xcal$ be a non-empty set sometimes referred
to as the index set, and $k$ a symmetric
real-valued\footnote{kernels are usually complex valued in the mathematical literature; we only consider the real case here, which is the common practice in machine learning.} function on
$\Xcal\times\Xcal$. For practitioners of kernel methods, a kernel is
above all a
positive definite function in the following sense:
\begin{definition}[Real-valued Positive Definite Kernels]
A symmetric function $k:\Xcal\times\Xcal\rightarrow \RR$ is a
positive definite (p.d.) kernel on $\Xcal$ if
\begin{equation}\label{eq:kernelnonneg}
\sum_{i,j=1}^n c_i c_j\,k\,(x_i,x_j) \geq 0,
\end{equation}
holds for any $n \in \NN , x_1,\ldots,x_n \in \Xcal$ and $c_1
\ldots,c_n \in \RR$.\label{def:positivedef}
\end{definition}

 \paragraph{kernel matrices derived from kernel functions:}
one can easily deduce from Definition~\ref{def:positivedef} that
the set of
p.d.~kernels is a closed, convex pointed cone\footnote{A set $C$
is a cone if for any $\lambda\geq0, x\in C \Rightarrow \lambda x\in
C$, pointed if $x\in C,-x\in C\Rightarrow x=0$}. Furthermore, the
positive definiteness of kernel functions translates in practice
into the positive definiteness of so called \emph{Gram} matrices, that is
matrices of kernel evaluations built on a sample of points
$X=\{x_i\}_{i\in I}$ in $\Xcal$,
$$
K_X = \left[k(x_i,x_j)\right]_{i,j\in I}.
$$
Elementary properties of the set of kernel functions such as its
closure under pointwise and tensor products are directly inherited
from well known results in Kronecker and Schur (or Hadamard)
algebras of matrices~\citep[\S 7]{Bernstein}. 

\paragraph{kernel matrices created using other kernel matrices:}
kernel matrices for a sample $X$ can be obtained by applying
transformations $r$ that conserve positive definiteness to a prior
Gram matrix $K_X$. In such a case the matrix $r(K_X)$ can be used directly on that subspace, namely
without having to define explicit formulas for the constructed
kernel on the whole space $\Xcal\times\Xcal$. A basic example is known as the empirical kernel map, where the square map $r:M\rightarrow M^2$ can be used on a matrix~\citep{SchWesEskLesNob02}. More complex constructions are the computation of the diffusion kernel on
elements of a graph through its Laplacian matrix~\citep{ kondor02},
or direct transformations of the kernel matrix through unlabelled
data~\citep{Sindhwani}.

\paragraph{strict and semi-definite positiveness:}
functions for which the sum in Equation~\eqref{eq:kernelnonneg} is
(strictly) positive when $c\neq 0$ are sometimes referred to as
positive definite functions, in contrast with functions for which
this sum is only non-negative, which are termed positive
\emph{semi}-definite. We will use for convenience throughout this
report the term positive definite for kernels that simply comply
with non-negativity, and will consider indifferently positive
semi-definite and positive definite functions. Most theoretical
results that will be presented in this report are also indifferent to this
distinction, and in numerical practice definiteness and
semi-definiteness will be equivalent since most estimation procedures consider a regularization of some form on the matrices to explicitly lower bound their conditioning number\footnote{that is the ratio of the biggest to the smallest eigenvalue of a matrix}.

\paragraph{the importance of positive definiteness :} 
Equation~\eqref{eq:kernelnonneg} distinguishes general measures of similarity between objects and a kernel function. The requirement of Equation~\eqref{eq:kernelnonneg} is important when seen from (at least) two perspective. First, the usage of positive definite matrices is a key assumption in convex programming~\cite{Boyd:1072}. In practice the positive definiteness of kernel matrices ensures that kernel algorithms such as Gaussian processes or support vector machines converge to a relevant solution\footnote{~\citep{pami_Haasdonk05,alexnips2007} show however that arbitrary similarity measures can be used with slightly modified kernel algorithms } 
Second, the positive definiteness assumption is also a key assumption of the functional view described below in reproducing kernel
Hilbert spaces theory.

\subsection{Reproducing Kernels}\label{subsubsec:reproducingkernel}
Kernels can be also viewed from the functional analysis viewpoint, since to each kernel $k$ on $\Xcal$ is associated a Hilbert space
$\Hcalk$ of real-valued functions on $\Xcal$.
\begin{definition}[Reproducing Kernel]\label{def:reprokernel}
A real-valued function $k:\Xcal\times\Xcal\rightarrow \RR$ is a reproducing
kernel of a Hilbert space $\Hcal$ of real-valued functions on
$\Xcal$ if and only if
$$
\begin{aligned}
\text{i)}&\;\forall t \in \Xcal, \;\;k(\cdot,t)\in\Hcal;\\
\text{ii)}&\;\forall t \in \Xcal, \forall f\in\Hcal,
 \;\;\langle f, k(\cdot,t) \rangle = f(t).
\end{aligned}
$$
\end{definition}
Condition (ii) above is called the \emph{reproducing property}.
A Hilbert space that is endowed with such a kernel is called a
reproducing kernel Hilbert space (rkHs) or a proper Hilbert space.
Conversely, a function on $\Xcal\times\Xcal$ for which such a
Hilbert space $\Hcal$ exists is a reproducing kernel and we
usually write $\Hcalk$ for this space which is unique. It turns
out that both Definitions~\ref{def:positivedef}
and~\ref{def:reprokernel} are equivalent, a result known as the
Moore-Aronszajn theorem~\citep{aron50}. First, a reproducing
kernel is p.d., since it suffices to write the expansion of
Equation~\eqref{eq:kernelnonneg} to obtain the squared norm of the
function $\sum_{i=1}^{n}c_i k(x_i,\cdot)$, that is
\begin{equation}\label{eq:kernelnorm}
\sum_{i,j=1}^n c_i c_j\,k\,(x_i,x_j)= \left\|\sum_{i=1}^{n}c_i
k(x_i,\cdot)\right\|_{\Hcal}^2,
\end{equation}
which is non-negative. To prove the opposite in a general setting,
that is not limited to the case where $\Xcal$ is compact which is
the starting hypothesis of the Mercer representation
theorem~\citep{Mercer:09} reported in~\citep[p.37]{schoelkopf02learning}, we refer the reader to the
progressive construction of the rkHs associated with a kernel $k$
and its index set $\Xcal$ presented in~\citep[\S 1.3]{berlinet03reproducing}. In practice, the rkHs boils down to
the completed linear span of elementary functions indexed by
$\Xcal$, that is
$$
\Hcal_k\defeq \overline{\spa}\{k(x,\cdot), x\in\Xcal\},
$$
whereby completeness we mean that all Cauchy sequences of functions converge.

\paragraph{the parallel between a kernel and a rkHs:} Definition~\ref{def:reprokernel} may seem theoretical at first glance, but its consequences are are however very practical. Defining a positive
definite kernel $k$ on any set $\Xcal$ suffices to inherit a
Hilbert space of functions $\Hcalk$ which may be used to pick candidate functions for a given data-analysis task. By selecting a kernel $k$, we
hope that the space $\Hcalk$ -- though made up of \emph{linear}
combinations of elementary functions -- may contain useful
functions with low norm. This is in many ways equivalent to defining a space of low degree polynomials and its dot-product in order to approximate an arbitrary function of interest on a given interval $[a,b]$ on the real line with a polynomial of low norm.

\paragraph{functional norm in a rkHs:} another crucial aspect of rkHs is the simplicity of their induced norms
and dot-products which are both inherited from the reproducing
kernel. The fact that this norm is easy to compute for finite
expansions, as seen in Equation~\eqref{eq:kernelnorm}, is an important property
which has direct implications when considering regularized estimation schemes, introduced in Section~\ref{sec:supervized} and more precisely Equation~\eqref{eq:regrisk}. Additionally, the
dot-product between two functions in the rkHs can be expressed as 
$$
\left\langle \sum_{i\in I} a_i k(x_i,\cdot) , \sum_{j\in J} b_j
k(y_j,\cdot)\right\rangle = \sum_{i\in I, j\in J} a_i b_j
k(x_i,y_j).
$$
which only depend on kernel evaluations on pairs $(x_i,y_j)$ and on the weights $a_i$ and $b_j$. The fact that in $\Hcalk$ the dot-product $\langle
k(x,\cdot),k(y,\cdot)\rangle_{\Hcalk}$ is equal to $k(x,y)$
illustrates an alternative view, namely that a kernel is a disguised dot-product.
\subsection{Kernels as Feature Maps}
The theorem below~\citep[p.22]{berlinet03reproducing} gives an
interpretation of kernel functions, seen as dot-products between
feature representations of their arguments in a space of
sequences.
\begin{theorem}
A function $k$ on $\Xcal\times\Xcal$ is a positive definite kernel
if and only if there exists a set $T$ and a mapping $\phi$ from
$\Xcal$ to $l^2(T)$, the set of real sequences $\{u_t, t\in T\}$
such that $\sum_{t\in T}|u_t|^2<\infty$, where
$$
\forall (x,y\,)\in \Xcal \times \Xcal\,,\, k(x,y)= \sum_{t\in T}
\phi\,(x)_t \phi\,(y)_t=\langle \phi(x),\phi(y)\rangle_{l^2(X)}
$$
\end{theorem}
The proof is derived from the fact that for any Hilbert space
(notably $\Hcalk$) there exists a space $l^2(X)$ to which it is
isometric. As can be glimpsed from this sketch, the feature map
viewpoint and the rkHs one are somehow redundant, since
$$
x\mapsto k(x,\cdot),
$$
is a feature map by itself. If the rkHs is of finite dimension,
functions in the rkHs are exactly the dual space of the
Euclidian space of feature projections. Although closely
connected, it is rather the feature map viewpoint than the rkHs
one which actually spurred most of the initial advocation for
kernel methods in machine learning, notably the SVM as presented
in~\citep{mach:Cortes+Vapnik:1995,schoelkopf02learning}. The
the latter references present kernel machines as mapping data-entries into high-dimensional feature spaces,
$$
\{x_1,\cdots,x_n\} \mapsto \{\phi(x_1),\cdots,\phi(x_n)\},
$$
to find a linear decision surface to separate the points in two
distinct classes of interest. This interpretation actually
coincided with the practical choice of using polynomial
kernels\footnote{$k(x,y)=(\langle x,y\rangle +b)^d, d\in\NN,
b\in\RR^+$} on vectors, for which the feature space is of finite
dimension and well understood as products of monomials up to degree $d$.

The feature map approach was progressively considered
to be restrictive in the literature, since it imposes to consider first the
extracted features and then compute the kernel that matches them.
Furthermore, useful kernels obtained directly from a similarity between
objects do not always translate into feature maps which can be easily described, as
in diffusion kernels on graphs for instance~\citep{kondor02}. Kernels without explicit feature maps may also be
obtained through the polynomial combination of several kernels. The feature map formulation, particularly
advocated in the early days of SVM's, also misled some observers
into thinking that the kernel mapping was but a piece of the SVM
machinery. Instead, the SVM should be rather seen as an efficient
computational approach -- among many others -- deployed to select
a ``good'' function $f$ in the rkHs $\Hcalk$ given a learning sample, as
presented in Section~\ref{sec:supervized}.

\subsection{Kernels and Distances, a Discussion}\label{sec:distndpd}
We discuss in this section possible parallels between positive definite kernels and distances. Kernel methods are often
compared to distance based methods such as
nearest neighbors. We would like to point out a few differences between their two respective ingredients, kernels $k$ and distances $d$.
\begin{definition}[Distances] Given a  space $\Xcal$, a nonnegative-valued function $d$ on $\Xcal\times\Xcal$ is a distance if it satisfies the following axioms, valid for all elements $x,y$ and $z$ of $\Xcal$:
\begin{itemize}
\item $d(x,y) \leq 0$, and $d(x,y)$ = 0 if and only if x = y.
\item $d(x,y) = d(y,x)$ (symmetry),
\item $d(x,z) \geq d(x,y) + d(y,z)$ (triangle inequality)
\end{itemize}
\end{definition}
the missing link between kernels and distances is given by a particular type of kernel function, which includes all negations of positive definite kernels as a particular case,
\begin{definition}[Negative Definite Kernels]
A symmetric function $\psi:\Xcal\times\Xcal\rightarrow \RR$ is a
negative definite (n.d.) kernel on $\Xcal$ if
\begin{equation}\label{eq:kernelnegdef}
\sum_{i,j=1}^n c_i c_j\,\psi\,(x_i,x_j) \leq 0
\end{equation}
holds for any $n \in \NN , x_1,\ldots,x_n \in \Xcal$ and $c_1
\ldots,c_n \in \RR$ such that $\sum_{i=1}^{n}c_i=0$.
\end{definition}

A matricial interpretation of this is that for any set of points $x_1,\cdots,x_n$ and vectors of weights $c\in\RR^n$ in the hyperplane 
$\left\{y\; |\; 1^T y=0\right\}$.
we necessarily have that  $c^T\Psi c\leq 0$ with $\Psi=[\psi(x_i,x_j)]_{i,j}$. A particular family of distances known as Hilbertian norms can be considered as negative definite
kernels as pointed out in~\cite{hein05hilbertian}. This link is made explicitly by~\cite[Proposition 3.2]{berg84harmonic} given below
\begin{proposition}
Let $\Xcal$ be a nonempty set and $\psi:\Xcal\times\Xcal$ be a negative definite kernel. Then there is a Hilbert space $H$ and a mapping $x\mapsto\phi(x)$ from $X$ to $H$ such that 
\begin{equation}\label{eq:negdef}
\psi(x,y)= \|\phi(x)-\phi(y)\|^2+f(x)+f(y) ,
\end{equation}
where $f:\Xcal\rightarrow \RR$ is a real-valued complex function on $X$. If $\psi(x,x)=0$ for all $x\in\Xcal$ then $f$ can be chosen as zero. If the set of pairs such that $\psi(x,y)=0$ is exactly $\{(x,x),x\in\Xcal\}$ then $\sqrt{\psi}$ is a distance.
\end{proposition}

\paragraph{negative definite kernels and distances:} the parallel between negative definite kernels and distances is thus clear: whenever a n.d. kernel vanishes on the \emph{diagonal}, that is the set $\{(x,x),x\in\Xcal\}$, and is zero only on the diagonal, then its square root is a distance for $\Xcal$. More generally, to each negative definite kernel corresponds a decomposition~\eqref{eq:negdef} which can be exploited to recover a distance given that the function $f$ can be deduced from $\psi$, typically as $\frac{\psi(x,x)}{2}$. On the other hand, to each distance does not correspond necessarily a negative definite kernel and there are numerous examples of distances which are not Hilbertian metric such as the Monge-Kantorovich distance~\citep{naor-2005} or most variations of the edit distance~\citep{VerSaiAku04}.

\paragraph{negative definite kernels and positive definite kernels:} on the other hand, n.d. kernels can be identified with a subfamily of p.d. kernels known as infinitely divisible kernels. A nonnegative-valued kernels $k$ is said to be infinitely divisible if for every $n\in\NN$ there exists a positive definite kernel $k_n$ such that $(k_n)^{1/n}$ is positive definite. 
\begin{example} A simple example is the usual Gaussian kernel between two vectors of $\RR^d$ since rewriting it as
$$
k_\sigma(x,y)=e^{-\frac{\|x-y\|^2}{2\sigma^2}}=\left(e^{-\frac{\|x-y\|^2}{2n\sigma^2}}\right)^n,
$$ 
suffices to prove this property. 
\end{example} 
Here follows a slightly simplified version of \cite[Proposition 2.7]{berg84harmonic} which provides a key interpretation:
\begin{proposition}\label{prop:inftydiv} For a p.d. kernel $k\geq 0$ on $\Xcal\times\Xcal$, the following conditions are equivalent
	\begin{enumerate}[(i)]
	\item $k$ is infinitely divisible,
	\item $-\log k\in\Ncal(\Xcal)$,
	\item $k^t$ is positive definite for al $t>0$.
\end{enumerate}

\end{proposition}

\begin{figure}[htbp]
\scalebox{.6}{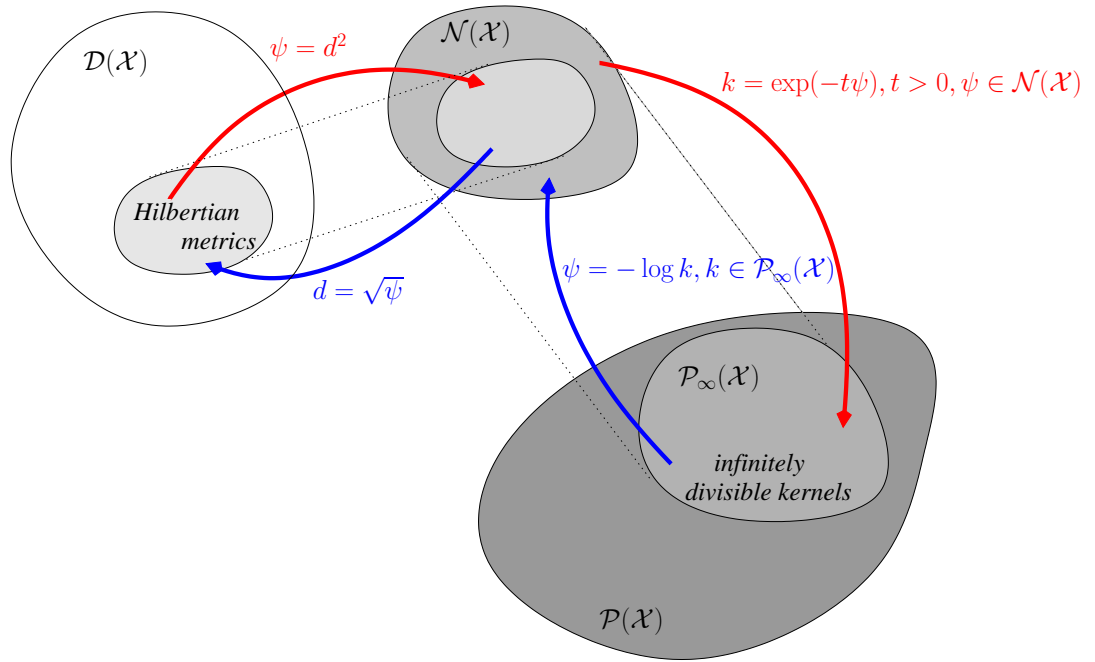}
\caption{A schematic view of the relationships between the set of distances on $\Xcal$, written as $\Dcal(\Xcal)$ on the left, its subset of Hilbertian metrics and their one-to-one mapping with a certain family of negative-definite kernels vanishing on the diagonal, itself contained in the set $\Ncal(\Xcal)$ of more general negative definite kernels on $\Xcal$. Note that the set of negative-definite kernels is in direct correspondence with the set $\Pcal_\infty(\Xcal)$, the subset of infinitely divisible positive definite kernels of $\Pcal(\Xcal)$.}\label{fig:distndpd}
\end{figure}
Figure~\ref{fig:distndpd} provides a schematic view on the relationships between distances, negative definite kernels and positive definite kernels. The reader should also keep in mind that 
\begin{enumerate}[(i)]
	\item $\Dcal(\Xcal)$ is a cone;
	\item $\Ncal(\Xcal)$ is a cone. Additionally, if $\psi(x,x)\geq 0$ for all $x\in\Xcal$, $\psi^\alpha$ and $\log(1+\psi)$ are also in $\Ncal(\Xcal)$ for $0<\alpha<1$ following~\cite[Corollary 2.10]{berg84harmonic};
	\item $\Pcal(\Xcal)$ is a cone. Additionally, $\psi^n$ is also in $\Pcal(\Xcal)$ for $n\in\NN$, as well as the pointwise product $\psi_1\psi_2$ of two p.d. kernels;
\end{enumerate}
	and may refer to~\cite[p.79]{berg84harmonic} for other numerous results and examples.

\newpage\section{Designing Kernels from Statistical Knowledge}\label{chap:creatingkernels}
We follow the relatively theoretical exposition of the previous chapter with a more practical exposition. Although the mathematical elements presented above explain most of the desirable properties of kernel machines, notably the convexity of most optimization carried out when estimating kernel machines, the view taken by practitioners on kernels is most often linked with that of a \emph{similarity} measure between objects. Namely that for two objects $x$ and $y$ the value $k(x,y)$ can be a reliable quantification of how $x$ and $y$ are similar. This similarity may be chosen arbitrarily, incorporating as much prior knowledge on the objects as possible without any connection to the task itself, or rather considered under the light of a given task.

\begin{example} The usual criterion for two texts to be similar might be that they share the same languages and/or topics of interest and/or overall length, but a very specialized algorithm might solely focus on the occurrence of a single word within their body. Two photographs might be qualified as similar if they display similar colors or shapes. For other tasks, rather than the image itself their date or their location might be the key criterion, regardless of their pictorial content. Two videos might be qualified as similar if they display the same person for a fixed length of time or if they were broadcasted through the same television channel.
\end{example}
The following sections start with the relatively simple example of defining kernels on vectors. We address objects with more complex structures later in this chapter using statistical modeling.

\subsection{Classical Kernels for Vectors}
\subsubsection{Vectors in $\RR^n$}\label{sec:vectorsrn}
Finite dimensional Vectors are a fundamental tool to represent natural phenomenon as numeric data. Vectors are known to be easy to manipulate by both algorithms and computer codes, and as such positive definite kernels taking vector arguments can be easily constructed. The canonical dot-product on a vector-space of finite dimensions, also known as the linear kernel $k(x,y)=x\cdot y$ is the most fundamental example. We will use three lemmas to show the reader how
most classical kernels can be easily reconstructed through the linear kernel. For a family of
kernels $k_1,\ldots,k_n,\ldots$
\begin{itemize}
\item The sum $\sum_{i=1}{n}{\lambda_i k_i}$ is positive definite, given
$\lambda_1,\ldots,\lambda_n\geq0$
\item The product $k_1^{a_1}\cdots k_n^{a_n}$ is positive definite,
given $a_1,\ldots,a_n\in\NN$
\item The limit $k\defeq\lim_{n\rightarrow\infty}k_n$ is positive
definite if the limit exists.
\end{itemize}
Using these properties listed in~\citep{berg84harmonic}, we can reconstruct
\begin{itemize}
\item the polynomial kernel $k_p(x,y)=(x\cdot y +b)^d, b>0, d\in\NN$,
simply because $b>0$ is a p.d. kernel, and so is $(x\cdot y +b)$ as a
consequence of the first property, and so is $(x\cdot y +b)^d$ as a
consequence of the second.
\item the Gaussian kernel
$k_\sigma(x,y)=e^{-\frac{\|x-y\|^2}{2\sigma^2}}$ which can be rewritten
in the following form $$k_\sigma(x,y)=
\left[e^{-\frac{\|x\|^2}{2\sigma^2}}e^{-\frac{\|y\|^2}{2\sigma^2}}\right
] \cdot \left[\sum_{i=0}^{\infty} \frac{(x\cdot y)^i}{i!}\right]$$
    The term in the first brackets is trivially a kernel, and so is the
term in the second part as a limit of positive definite kernels.
\end{itemize}

\subsubsection{Vectors in $\RR^n_+$ and Histograms}\label{sec:vectors}
Histograms are frequently encountered in
applications of machine learning to real-life problems. Indeed, most natural phenomena produce visible data, which the practitioner is likely to count to describe reality. As a consequence, most observations are usually available
under the form of nonnegative vectors of counts, which, if normalized, yield
histograms of frequencies. Metrics or divergences for general probability
measures, the obvious generalization of histograms, is the object of study of information
geometry~\cite{amar01b}. However, as hinted in Section~\ref{sec:distndpd}, a proper understanding of metrics and divergences for a certain class of objects cannot be immediately applied to define positive definite kernels.
Indeed, the Kullback-Leibler divergence, which has a fundamental
importance in information geometry, cannot be used as such in kernel
methods as it is neither symmetric nor positive/negative definite.

\paragraph{elementary kernels on positive measures:} it
has been shown however in~\cite{hein05hilbertian} that the following family of
squared metrics, respectively the Jensen Divergence, the $\chi$-square, Total Variation and two variations of the Hellinger distance, are all negative-definite kernels:
$$
\begin{aligned}
\psi_{JD}(\theta,\theta')&=
h\left(\frac{\theta+\theta'}{2}\right)-\frac{h(\theta)+h(\theta')}{2},\\
\psi_{\chi^2}(\theta,\theta')&=\sum_i \frac{(\theta_i -
\theta'_i)^2}{\theta_i + \theta'_i}, \quad
\psi_{TV}(\theta,\theta')=\sum_i |\theta_i - \theta'_i|,\\
\psi_{H_2}(\theta,\theta')&=\sum_i |\sqrt{\theta_i} -
\sqrt{\theta'_i}|^2,\quad
\psi_{H_1}(\theta,\theta')=\sum_i |\sqrt{\theta_i} -
\sqrt{\theta'_i}|.\\
\end{aligned}
$$
As a consequence, these metrics can all be used to definite positive
definite kernels using Proposition~\ref{prop:inftydiv} and the following formula:
$$k(\theta,\theta')=e^{-\frac{1}{t}\psi},$$
with $t>0$. Although histograms appear frequently in the study of objects such as images, through histograms of colors, and texts, through bags-of-words representations, their usage alone is restrictive when studying objects that carry a finer structure. In such a case, probability distributions that are tailored to capture better the interdependencies between smaller components in those objects can be used to define kernels as presented in the next section.

\subsection{Statistical Modeling and Kernels}\label{chap:smk}

\paragraph{Fisher kernel:}
\citet{jaakkola-haussler-99} first thought of using generative
models to build kernels that would provide in turn the necessary inputs of \emph{discriminative} machines, that is kernel classifiers. Although the principle outlined in the next lines can be applied to different pairs of datatypes/generative models, we follow the original presentation of their paper which focused on sequences.
\citet{jaakkola-haussler-99} observed that the hidden Markov model (HMM), which is known to capture
efficiently the behaviour of amino-acid sequences can be
used as an efficient feature extractor. The authors did so by
defining for each considered sequence a vector of features derived from an estimated HMM model, namely the Fisher score.
Given a measurable space $(\Xcal,\Bcal,\nu)$ and a parametric
family of absolutely continuous measures of $\Xcal$ represented by
their densities $\{p_\theta,\theta\in\Theta\subset \RR^d\}$, the
Fisher kernel between two elements $x,y$ of $\Xcal$ is
$$
k_{\hat{\theta}}(x,y) = \left(\frac{\partial \ln
p_\theta(x)}{\partial \theta}\big|_{\hat\theta}\right)^T
J^{-1}_{\hat\theta} \left(\frac{\partial \ln p_\theta(y)}{\partial
\theta}\big|_{\hat\theta}\right),
$$
where $\hat\theta$ is a parameter selected beforehand to match the
whole training set, and $J_{\hat\theta}$ is the Fisher information
matrix computed in $\hat\theta$. The statistical model not only
acts as a \emph{feature extractor} through the score vectors, but also
defines the \emph{Mahalanobis metric} associated with these vectors through
$J_{\hat\theta}$. We introduce the following alternative formulation of the kernel quoted in~\cite{jaakkola99using} using the $\nabla_\theta$ notation which stands for the gradient of a function computed at $\theta$
\begin{equation}\label{eq:fisherker}
k_{\hat{\theta}}(x,y) = e^{-\frac{1}{\sigma^2}\left(\nabla_{\hat\theta}\ln p_\theta(x)-\nabla_{\hat\theta}\ln p_\theta(y) \right)^T
J^{-1}_{\hat\theta} \left(\nabla_{\hat\theta}\ln p_\theta(x)-\nabla_{\hat\theta}\ln p_\theta(y) \right)},
\end{equation}

\paragraph{extensions to the Fisher kernel:}
 the proposal of the Fisher kernel fostered further research, notably
in~\citep{Tsuda:967,nips02-SP02}. The motivation behind these
contributions was to overcome the limiting assumption that the
parameter $\hat\theta$ on which the score vectors are evaluated is
unique and fits the whole set of points at hand. Rather,
~\citet{Tsuda:967} and~\citet{nips02-SP02} proposed simultaneously
to incorporate in the context of binary classification two
parameters $\hat\theta_1$ and $\hat\theta_2$ for each class
respectively, and consider the score vector of the likelihood
ratio between the two classes evaluated in $x$,
$$
\phi_{\hat\theta_1,\hat\theta_2}:x\mapsto \left(\frac{\partial \ln
\frac{p_{\theta_1}(x)}{p_{\theta_2}(x)}}{\partial
\vartheta}\Big|_{\,\hat\vartheta=(\hat\theta_1,\hat\theta_2)}\right),
$$
where $\vartheta=(\theta_1,\theta_2)$ is in $\Theta^2$, to
propose instead the kernel
$$
(x,y)\mapsto \phi_{\hat\theta_1,\hat\theta_2}(x)^T
\phi_{\hat\theta_1,\hat\theta_2}(y).
$$
The Fisher kernel was also studied from a theoretical perspective when used in conjunction 
with a logistic regression~\citep{Tsuda:966}.

\paragraph{mutual information kernels:}
the Fisher kernel is related to a wider class of kernels coined
down as mutual information kernels by~\citet{seeger02covariance}.
Starting also from a set of distributions $\{p_\theta,\theta\in
\Theta\}$ where $\Theta$ is measurable, and from a given prior
$\omega\in L_2(\Theta)$, the mutual information kernel $k_\omega$
between two elements $x$ and $y$ is defined as
\begin{equation}\label{eq:mik}
	k_\omega(x,y) = \int_{\Theta} p_\theta(x) p_\theta(y) \,\omega(d
\theta).
\end{equation}
As noted in~\citep{seeger02covariance}, the Fisher kernel can be
regarded as a maximum \textit{a posteriori} approximation of the
mutual information kernel, by setting the prior $\omega$ to the
multivariate Gaussian density
$\Ncal(\hat\theta,J^{-1}_{\hat\theta})$, following the
approximation of Laplace's method. Let us review this claim in more details:
 given an object $x$ and a parameter $\hat\theta$, the following approximation
$$\log p_\theta(x) \approx \log p_{\hat\theta}(x)+\nabla_{\hat\theta}\ln p_\theta(x)^T(\theta-\hat\theta)$$ 
can be rewritten using the notation $$
\Phi(x)=\nabla_{\hat\theta}\ln p_\theta(x)= \frac{\partial \ln
p_\theta(x)}{\partial \theta}\big|_{\hat\theta}
$$ as
$$\log p_\theta(x) \approx \log p_{\hat\theta}(x)+ \Phi(x)(\theta-\hat\theta).$$
Using a Gaussian approximation for $\omega$ yields a change in Equation~\ref{eq:mik} as
\begin{equation}\label{eq:mik2}
	\begin{aligned}
	k(x,y) &= C \int_{\Theta} e^{\log p_{\hat{\theta}}(x)+ \Phi(x)^T(\theta-\hat\theta) + \log p_{\hat\theta}(y)+ 
	\Phi(y)^T(\theta-\hat{\theta})}\;\;e^{-(\theta-\hat{\theta})^T J_{\hat{\theta}} (\theta-\hat{\theta})}d\theta\\
	& = C p_{\hat{\theta}}(x) p_{\hat{\theta}}(y) \int_{\Theta} e^{\left(\Phi(x)+\Phi(y)\right)^T(\theta-\hat\theta) + (\theta-\hat\theta)^T J_{\hat\theta} (\theta-\hat\theta)}d\theta\\
	& = C' p_{\hat{\theta}}(x) p_{\hat{\theta}}(y) e^{\frac{1}{2}(\Phi(x)+\Phi(y))^T J_{\hat\theta}^{-1} (\Phi(x)+\Phi(y))}
\end{aligned}
\end{equation}
it is then easy to check that the kernel
$$
\tilde{k}(x,y)=\frac{k(x,y)}{\sqrt{k(x,x)k(y,y)}}
$$
is equal to the Fisher kernel given in its form of Equation~\eqref{eq:fisherker}.
 ~\citet{cuturi05context} propose an example of a mutual information kernel defined on strings that can be computed exactly. In the latter work the set of distributions $\{p_\theta,\theta\in \Theta\}$ is a set of Markov chain
densities on sequences with finite depths. The prior $\omega$ is a combination of branching
process priors for the structure of the chain and mixtures of
Dirichlet priors for the transition parameters. This setting
yields closed computational formulas for the kernel through previous work led in universal coding~\citep{willems95contexttree,stflour}. The computations can be carried in a number of elementary operations that is linear in the lengths of the inputs $x$ and $y$.

\paragraph{marginalized kernels:}
in the framework of sequence analysis
first~\citep{tsuda02marginalized}, and then in comparisons of
graphs~\citep{KasTsuIno03}, further attention was given to latent
variable models to define kernels in a way that also generalized
the Fisher kernel. In a latent variable model, the probability of
emission of an element $x$ is conditioned by an unobserved latent
variable $s\in \Scal$, where $\Scal$ is a finite space of possible
states. When a string is considered under the light of a hidden
Markov model, to its chain $x=x_1\cdots x_n$ of letters is
associated a similar sequence $s=s_1\cdots s_n$ of states that is
not usually observed. When the sequence of states $s$ is known,
the probability of $x$ under such a model is then determined by
the marginal probabilities $p(x_i|s_i)$. Building adequate
transition structures for the emitting states, and their
corresponding emission probabilities is one of the goals of HMM
estimations. The marginalized kernel assumes that this sequence is
not known for objects $x$ and $y$, but it performs, given an
available structure of states, an averaging
$$
k(x,y) = \sum_{s\in\Scal} \sum_{s'\in \Scal} p(s|x) \,p(s'|y)
\,\kappa\left(\left(x,s\right),\left(y,s'\right)\right)
$$
of arbitrary kernel evaluations $\kappa$ weighted by posterior
probabilities which are estimated from data. In this setting,
$\kappa$ can be any arbitrary kernel on $\Xcal\times\Scal$. For
particular choices of $\kappa$ the kernel can be computed in
closed form, both on sequences and
graphs~\citep{Mahe2004Extensions}.

\paragraph{kernels defined on maximum-likelihood parameters:}
the previous approaches make different uses of a statistical model $p_\theta(x)$. In mutual information kernels $p_\theta(x)$ is treated as a feature indexed by a large set of parameters $\theta\in\Theta$. For marginalized kernels an unseen, latent variable is added to $p_\theta$, $p_\theta(x,s)$ and while $\theta$ is kept constant the integration of $p_\theta(x,s)p_\theta(y,s')$ is led over all possible combinations of latent variables $(s,s')$. A third approach, conceptually simpler, compares two objects by considering
directly the parameters $\theta$ and $\theta'$ that fits them better respectively, that is, map first
$$
(x,y)\mapsto (\hat\theta_x,\hat\theta_y)\in\Theta^2,
$$
through maximum likelihood estimation for instance, and then
compare $x$ and $y$ through a kernel $k_\Theta$ on $\Theta$,
$$
k(x,y)=k_\Theta(\hat\theta_x,\hat\theta_y).
$$
Under this form, the topic of defining interesting functions $k$ on $(\hat\theta_x,\hat\theta_y)$ is loosely connected with information geometry~\citep{amar01b} and one may use for simple densities some of the kernels presented in Section~\ref{sec:vectors}. For more complex spaces of parameters $\Theta$ one may refer to
~\citet{jebara04probability} which presents the family of kernels
$$
k_\beta(x,y)=\int_{\Xcal}p_{\hat\theta_x}(z)^\beta
p_{\hat\theta_y}(z)^\beta dz
$$
for $\beta>0$, the case $\beta=\frac{1}{2}$ being the well known
Bhattacharrya affinity between densities. The authors review a
large family of statistical models for which these kernels can be
computed in closed form, ranging from graphical models, Gaussian
multivariate densities, multinomials and hidden Markov models. 

\paragraph{information diffusion kernel:} aiming also at computing kernels of
interest on multinomials,~\citet{Lafferty2005} propose to follow~\citet{ kondor02}
and use diffusion processes to define kernels. To do so they
express solutions for the heat equation in the Riemannian manifold
induced by the Fisher metric of the considered statistical models,
inspired again by information geometry~\citet{amar01b}. They derive \emph{information diffusion}
kernels out of such solutions which, when specialized to
multinomials, that is elements of the simplex\footnote{writing
$\Sigma_d$ for the canonical simplex of dimension $d$, i.e.,
$\Sigma_d=\{\xi=(\xi_i)_{1\leq i \leq d} : \xi_i\geq 0, \sum
\xi_i=1\}$.}, boil down to kernels of the form
\begin{equation}\label{eq:geod}
k_{\Sigma_d}(\theta,\theta')=e^{-\frac{1}{t}\arccos^2(\sqrt{\theta\cdot\
\theta'})},
\end{equation}
where $t>0$ is the diffusion parameter. Note that the squared
arc-cosine in Equation~\eqref{eq:geod} is the squared geodesic
distance between $\theta$ and $\theta'$ seen as elements from the
unit sphere (that is when each $\theta_i$ is mapped to
$\sqrt{\theta_i}$). Based on the seminal work
of~\citet{Scho:1942},~\citet{zhang2005} rather advocate the direct
use of the geodesic distance:
$$
k_{\Sigma_d}(\theta,\theta')=e^{-\frac{1}{t}\arccos(\sqrt{\theta\cdot\
\theta'})},
$$ 
They prove that the geodesic distance is a negative definite kernel on
the \emph{whole sphere}, while its square used in
Equation~\eqref{eq:geod} is not. If the points $\theta$ and $\theta'$ are restricted to lie in the positive
orthant, which is the case for multinomials, both approaches yield
however positive definite kernels.

\subsection{Semigroup Kernels and Integral Representations}
Most positive definite kernels on groups, which includes kernels on vectors of $\RR^n$ as described in Section~\eqref{sec:vectorsrn} can be considered as semigroup kernels. A semigroup is an algebraic structure that is simple enough to fit most datatypes and rich enough to allow for a precise study of the kernels defined on them. Most of the material of this section is taken from~\citep{berg84harmonic}, but the interest reader may consult the additional references~\citep{devinatz1955representation,ehm2003stationary}. Let us start this section with the following definitions.

\paragraph{semigroups:}  a semigroup $(\Scal,+)$ is a nonempty set $\Scal$ endowed with an
\emph{associative composition} $+$ which admits a
neutral element $0$, that is such that $\forall x \in \Scal, x+0=x$. An involutive semigroup $(\Scal,+,*)$ is a semigroup endowed with an involution $*$ which is a mapping $\Scal\rightarrow \Scal$ such that for any $x$ in $\Scal$ $(x^*)^*=x$. Let us provide some examples of semigroups:
\begin{itemize}
	\item $\Scal$ is the set of strings formed with letter from a given alphabet, $+$ is the concatenation operation, $0$ is the empty string and $*$ is either the identity or the operation which inverses the order of the letters of a string.
	\item $\Scal$ is a group, and $*$ is the inverse operation of the group. $(\RR,+,-)$ is a typical example.
	\item $\Scal$ is the positive orthant $\RR^+$ endowed with the usual addition and $*$ is the identity.
\end{itemize}
Note that most semigroups considered in the machine learning literature are \emph{abelian} that is operation $+$ is commutative.

\paragraph{semigroup kernels:} a semigroup kernel is a kernel $k$ defined through a complex-valued function $\varphi$ defined on $\Scal$ such that $$k(x,y)\overset{\defi}{=} \varphi(x+ y^*).$$ 
A function $\varphi$ is a positive definite function if the kernel that can be derived from it as $\varphi(x+y^*)$ is itself positive definite. When $\Scal$ is a vector space, and hence a group, for two elements $x,y$ of $\Scal$ one can easily check that most elementary kernels are either defined as
$$k(x,y)=\varphi(x - y),$$  or \begin{equation}\label{eq:autoinvo}k(x,y)=\psi(x + y),\end{equation} respectively when $*$ is the minus operation and $*$ is the identity. Kernels build on the former structure will typically emphasize the difference between two elements, given this difference can be computed, and include as their most important example radial basis functions (RBF) and the Gaussian kernel. When a subtraction between elements cannot be defined as is the case with strings, histograms and nonnegative measures, the form of Equation~\eqref{eq:autoinvo} is better suited as can be seen in some of the examples of Section~\ref{sec:vectors} and studied in~\citep{cuturi05semigroup}. In this work, the authors study a family of kernels for probability measures $\mu$ and $\mu'$ by looking at their average $(\mu+\mu')/2$. They narrow down their study to kernels defined through the variance matrix $\Sigma(\frac{\mu+\mu'}{2})$ of their average, and show that $$k(\mu,\mu')\defeq\frac{1}{\sqrt{\det\Sigma\left(\frac{\mu+\mu'}{2}\right)}},$$ is a positive definite kernel between the two measures. This result can be further extended through reproducing kernel Hilbert space theory, yielding a kernel between two clouds of points $\{x_1,\ldots,x_n\}$ and $\{y_1,\ldots,y_m\}$  which only depends on the kernel similarity matrices $K_{XY}=[\kappa(x_i,y_j)], K_X=[\kappa(x_i,x_j)]$ and $K_Y=[\kappa(y_i,y_j)]$.

\paragraph{integral representations:} semigroup kernels can be expressed as sums of semicharacters, a family of elementary functions on $\Scal$. A real-valued function $\rho$ on an Abelian semigroup $(S,+)$ is called a semicharacter if it satisfies  

\begin{enumerate}[(i)]
	\item $\rho(0) = 1,$
\item $\forall s, t \in\Scal, \;\rho(s + t) = \rho(s)\overline{\rho(t)},$
\item $\forall s \in\Scal, \;\rho(s)=\overline{\rho(s^*)}.$
\end{enumerate}
The set of semicharacters defined on $S$ is written $S^*$ while the set of bounded semicharacters can be written as $\hat{S}$. It is trivial to see that every semicharacter is itself a positive definite function. The converse is obviously not true, but it is possible to show that bounded semicharacters are the extremal points of the cone of bounded positive definite functions, therefore providing the following result given by~\cite{berg84harmonic}:
\begin{theorem}[Integral representation of p.d. functions]
A bounded function $\varphi: S\rightarrow \RR$ is p.d. if and only if it there exists a non-negative measure $\omega$ on $\hat{S}$ such that: 
$$\varphi(s)=\int_{\hat{S}}\rho(s)\; d\omega(\rho).$$ In that case the measure $\omega$ is unique.
\end{theorem}

When $S$ is the Euclidian space $\RR^d$ the following results due originally to Bochner and Bernstein respectively allow us to characterize kernels for two vectors $x$ and $y$ that depend respectively on $(x-y)$ and $(x+y)$.

\paragraph{identical involution} let a kernel $k$ be such that $k(x,y)=\varphi(x-y)$. Then there exists a unique non-negative measure $\omega$ on $\RR^d$ such that $$\varphi(x)=\int_{\RR^d} e^{i x^T r}d\omega(r);$$ In other words, $\varphi$ is the Fourier transform of a non-negative measure $\omega$ on $\RR^d$.

\paragraph{opposite involution} let a bounded kernel $k$ be such that $k(x,y)=\psi(x+y)$. Then there exists a unique non-negative measure $\omega$ on $\RR^d$ such that $$\psi(x)=\int_{\RR^d} e^{-x^T r}d\omega(r);$$ or in other words $\psi$ is the Laplace transform of a non-negative measure $\omega$ on $\RR^d$.

\newpage\section{Kernel Machines}\label{chap:machines}
Kernel machines are algorithms that select functions with desirable properties in a pre-defined reproducing kernel Hilbert space (rkHs) given sample data. All kernel estimation procedures define first a criterion that is a combination of possibly numerous and different properties. Subsequently, the element $f$ of the rkHs that is the optimum with respect to this criterion is selected following an optimization procedure. Before presenting such algorithms, let us mention an important theoretical challenge that appears when dealing with the estimation of functions in rkHs.

\par Let
$\Xcal$ be a set endowed with a kernel $k$ and $\Hcalk$ its
corresponding rkHs. Choosing a function in an infinite dimension space such $\Hcal$ can become an ill-defined problem when the criterion used to select the function does not have a unique minimizer. The representer theorem formulated below provides a practical answer to this problem when a regularization term is used along with a convex objective.
\subsection{The Representer Theorem}
\par Most estimation procedures presented in the statistical
literature to perform dimensionality reduction or infer a
decision function out of sampled points rely on the optimization
of a criterion which is usually carried out over a class of linear
functionals of the original data. Indeed, PCA, CCA, logistic regression and
least-square regression and its variants (lasso or ridge
regression) all look for linear transformations of the original
data points to address the learning task. When these optimizations
are led instead on an infinite dimensional space of functions,
namely in the rkHs $\Hcalk$, the optimization can be performed in
finite subspaces of $\Hcalk$ if the criterion only depends on a criterion computed on a finite sample of points. This
result is known as the representer theorem and explains why so
many linear algorithms can be ``kernelized'' when trained on
finite datasets.

\begin{theorem}[Representer
Theorem~\citep{Kimeldorf1971Some}]\label{theo:representer} Let
$\Xcal$ be a set endowed with a kernel $k$ and $\Hcalk$ its
corresponding rkHs. Let $\{x_i\}_{1\leq i\leq n}$ be a finite set
of points of $\Xcal$ and let $\Psi:\RR^{n+1}\rightarrow\RR$ be any
function that is strictly increasing with respect to its last
argument. Then any solution to the problem
$$
\min_{f\in\Hcalk}
\Psi\left(f(x_1),\cdots,f(x_n),\|f\|_{\Hcalk}\right)
$$
is in the finite dimensional subspace $\spa\{k(x_i,\cdot), 1\leq
i\leq n\}$ of $\Hcalk$.
\end{theorem}
The theorem in its original form was cast in a more particular
setting, where the term $\|f\|_{\Hcalk}$ would be simply added to
an empirical risk as often used in Section~\ref{sec:supervized}. This generalized version is however important to deal with an unsupervised setting.

\subsection{Eigenfunctions in a rkHs of Sample Data Points}\label{subsubsec:unsupervized}
In machine learning, unsupervised learning is a class of problems in which one seeks to determine how the data are organized. Indeed, for some applications practitioners are interested first in summarizing the information contained in their observations rather than inferring a decision function on such data. This task can be broadly categorized as dimensionality reduction and can be seen as a data-dependent way to summarize the information contained in each datapoint to a few numbers. Namely, given a sample $X=\{x_1,\cdots,x_n\}$ of points of $\Xcal$ translate such a set of points into an alternative representation $\xi=\{\xi_1,\cdots,\xi_n\}$ of such points where each $\xi_i$ is in $\RR^d$ and $d$ has a much lower dimensionality than $\Xcal$. 

If both $\Xcal$ and $\Ycal$ are Euclidian spaces, two popular unsupervized linear techniques are of particular interest.
\paragraph{Principal component analysis (PCA)} which aims at defining
an orthonormal basis $v_1,\cdots,v_{\dim(\Xcal)}$ of $\Xcal$ such
that for $1\leq j\leq\dim(\Xcal)$,
\begin{equation}\label{eq:critpca}
v_j=\underset{v\in\Xcal, \|v\|_\Xcal=1,
v\bot\{v_1,\cdots,v_{j-1}\}}{\argm}\var_X[\langle v, x
\rangle_{\Xcal}],
\end{equation}
where for any function $f:\Xcal\rightarrow \RR$, $\var_X[f]$
denotes the empirical variance with respect to the points
enumerated in $X$, that is $E_X[(f-E_X[f])^2]$. The $r$
first eigenvectors $v_1,\cdots,v_r$ are significative since the $r$ dimensional projection of $X$, $(v_1^T x, \cdots, v_r^T x)$ usually suffices to capture most of the variability of the data under a Gaussian assumption. 

\paragraph{Canonical correlation analysis (CCA)} can be applied when a set of measurements from a sample $X$ can be paired with another set of observations $Y=\{y_i\}_{1\leq i\leq n}$ taken in a set $\Ycal$, and that the pairs $(x_i,y_i)$ are drawn from a i.i.d law. Such tasks appear typically when each index $i$ refers to the same underlying object cast in different
modalities~\citep{vert03}. CCA looks for meaningful relationships between $X$ and $Y$ by focusing on linear projections of $X$ and $Y$, $\alpha^T X$ and $\beta^T Y$, such that the correlation between $\alpha^T X$ and $\beta^T Y$ is high. 
In mathematical terms this amounts to defining
\begin{equation}\label{eq:critcca}
\begin{aligned}
(\alpha,\beta)&= \underset{\xi\in\Xcal,\zeta\in\Ycal}{\argm} \corr_{X,Y} [\alpha^T, \beta^T]\\
&=\underset{\alpha\in\Xcal,\beta\in\Ycal}{\argm} \frac{\cov_{X,Y}[\alpha^T,\beta^T]}{\sqrt{\var_X[\alpha^T]\var_Y[\beta^T]}}
\end{aligned}
\end{equation}
where for two real valued functions $f:\Xcal\rightarrow \RR$ and
$g:\Ycal\rightarrow \RR$ we write
$$
\begin{aligned}
\var_{X}[f]&=E_{X}(f(x)-E_X[f(y)])^2,\\
\var_{Y}[g]&=E_{X}(g(y)-E_Y[g(y)])^2,\\
\cov_{X,Y}[f,g]&=E_{X,Y}[(f(x)-E_X[f(x)])(g(y)-E_Y[g(y)])].
\end{aligned}
$$

We observe that both optimizations look for vectors in $\Xcal$ as well as $\Ycal$ in the case of CCA that will be representative of the data dependencies. The three operators $\var_X$, $\var_Y$ and $\cov_{X,Y}$ can be approximated by finite sample estimators, respectively
$$
\begin{aligned}
\var^n_X[f]&=\sum_{i=1}^n \left(f(x_i)-\frac{1}{n}\sum_{j=1}^n f(x_j)\right)^2,\\
\var^n_Y[g]&=\sum_{i=1}^n \left(g(y_i)-\frac{1}{n}\sum_{j=1}^n g(y_j)\right)^2,\\
\cov^n_{X,Y}[f,g]&=\sum_{i=1}^n \left(f(x_i)-\frac{1}{n}\sum_{j=1}^n f(x_j)\right)\left(g(y_i)-\frac{1}{n}\sum_{j=1}^n g(y_j)\right).
\end{aligned}
$$

\paragraph{Generalization to functions in a rkHs:} The ``kernelization'' of such algorithms is natural when
considering the same criterions on the mappings in $\Hcal$ of the random variables $X$ and $Y$. We write for convenience $\Hcal_\Xcal$ and $\Hcal_\Ycal$ for the rkHs associated with $\Xcal$ and $\Ycal$
with respective kernels $k_\Xcal$ and $k_\Ycal$. If we cast now the problem as that of estimating a functions $f$ in $\Hcal_\Xcal$ and a couple of functions $(f,g)$ in $\Hcal_\Xcal$ and $\Hcal_\Ycal$ respectively, we
are now looking for vectors in such spaces -- that is real-valued functions on $\Xcal$, and $\Xcal\times\Ycal$ respectively -- that are directions of interest in the sense that they have adequate values according to the criterions defined in Equations~\eqref{eq:critpca} and~\eqref{eq:critcca}. When considered on
the finite subspaces of $\Hcalk$ spanned by the datapoints, the two previous optimizations
become
$$
f_j=\underset{f\in\Hcal_\Xcal, \|f\|_{\Hcal_\Xcal}=1,
f\bot\{f_1,\cdots,f_{j-1}\}}{\argm}\var_X[\langle f,
k_\Xcal(x,\cdot) \rangle_{\Hcal_\Xcal}],
$$
for $1\leq j\leq n$ and
\begin{equation}\label{eq:kcca}
(f,g)= \underset{f\in{\Hcal_\Xcal},g\in{\Hcal_\Ycal}}{\argm}
\frac{\cov_{X,Y} [\langle f,
k_\Xcal(x,\cdot)\rangle_{\Hcal_\Xcal}, \langle g,
k_\Ycal(y,\cdot)\rangle_{\Hcal_\Xcal}]}{\sqrt{\var_X[\langle f,
k_\Xcal(x,\cdot)\rangle_{\Hcal_\Xcal}]\var_Y[\langle g,
k_\Ycal(y,\cdot)\rangle_{\Hcal_\Ycal}]}}.
\end{equation}

\paragraph{kernel PCA:} the first problem has been termed kernel-PCA by~\citet{schoelkopf98kpca} and boils down to the decomposition
of the operator $\var^n_X$ into $n$ eigenfunctions\footnote{Note that kernelizing weighted PCA is not as straightforward and
can be only carried out through a more generalized eigendecomposition,
as briefly formulated in~\citep{cuturi04semigroup}}. This decomposition can be carried out by considering the $n\times n$ kernel matrix $K_X$ of the $n$ observations, or more precisely its centered counterpart 
$$\bar{K}_X=(I_n -\frac{1}{n}\mathds{1}_{n,n})K_X(I_n -\frac{1}{n}\mathds{1}_{n,n}).$$
The eigenfunctions $f_i$ can be recovered by considering the eigenvalue/eigenvector pairs $(e_i,d_i)$ of $\bar{K}_X$, that is such that
$$
\bar{K}_X = E D E^T
$$ 
where $D=\diag(d)$ and $E$ is an orthogonal matrix. Writing $U= ED^{-1/2}$ we have that
\begin{equation}\label{eq:kpcadef}
f_j(\cdot) = \sum_{i=1}^n U_{i,j} k(x_i,\cdot)
\end{equation}
with $\var^n_X[f_j(x)]=\frac{d_j}{n}$.

\paragraph{kernel CCA:} the second optimization, first coined down
kernel-CCA by~\citet{Akaho_2001_kcca}, is ill-posed if
Equation~\eqref{eq:kcca} is used directly with a finite sample, and requires a
regularization as explained in~\citep{fbach-kernel,fukumizu-kcca}. Namely, the direct maximization
$$ 
(f,g)= \underset{f\in\Xcal,g\in\Ycal}{\argm} \frac{\corr^n_{X,Y} [f, g]}{\sqrt{\var^n_X[f]\var^n_Y[g]}}
$$
is likely to result in degenerated directions where $\var^n_X[f]$ or $\var^n_Y[g]$ is close to zero, which suffices to maximize the ratio above. Instead, the criterion below,
$$
(f,g)= \underset{f\in\Xcal,g\in\Ycal}{\argm} \frac{\corr^n_{X,Y} [f, g]}{\sqrt{(\var^n_X[f]+\lambda\|f\|^2)(\var^n_Y[g]+\lambda\|g\|^2)}},
$$
is known to converge to a meaningful solution when $\lambda$ decreases to zero as $n$ grows with the proper convergence speed~\citep{fukumizu-kcca}. The finite sample estimates $f^n$ and $g^n$ can be recovered as
$$
\begin{aligned}
f^n(\cdot) &= \sum_{i=1}^n \xi_i \varphi_i(\cdot),\\
g^n(\cdot) &= \sum_{i=1}^n \zeta_i \psi_i(\cdot)
\end{aligned}
$$
where $\xi$ and $\zeta$ are the solutions of 
$$
(\xi,\zeta)=\underset{\xi,\zeta\in \RR^n,\\ \xi^T(\bar{K}_X^2+n\lambda \bar{K}_X)\xi=\zeta^T(\bar{K}_Y^2+n\lambda \bar{K}_Y)\zeta=1}{\argm}{\zeta^T \bar{K}_Y\bar{K}_X \xi}
$$
and 
$$
\begin{aligned}
\varphi_i(\cdot)=k_\Xcal(x_i,\cdot)-\frac{1}{n}\sum_{j=1}^n k_\Xcal(x_i,\cdot),\\
\psi_i(\cdot)=k_\Ycal(y_i,\cdot)-\frac{1}{n}\sum_{j=1}^n k_\Ycal(y_i,\cdot),
\end{aligned}
$$
are the centered projections of $(x_i)$ and $(y_j)$ in $\Hcal_\Xcal$ and $\Hcal_\Ycal$ respectively. The topic of supervised dimensionality reduction, explored
in~\citep{fukumizu-kernel}, is also linked to the kernel-CCA
approach. The author look for a sparse representation of the data
that will select an effective subspace for $\Xcal$ and delete all directions in $\Xcal$ that are not correlated to paired observations in $\Ycal$, based on two samples $X$ and $Y$. In linear
terms, such a sparse representation can be described as a
projection of the points of $\Xcal$ into a subspace of lower
dimension while conserving the correlations observed with
corresponding points in $\Ycal$.

\subsection{Regression, Classification and other Supervised Tasks}\label{sec:supervized}
Suppose that we wish to infer now from what is observed in the
samples $X$ and $Y$ a causal relation between all the points of
$\Xcal$ and $\Ycal$. This type of inference is usually restricted
to finding a mapping $f$ from $\Xcal$ to $\Ycal$ that is
consistent with the collected data and has desirable smoothness
properties so that it appears as a ``natural'' decision function
seen from a prior perspective. If $\Xcal$ is Euclidian and $\Ycal$
is $\RR$, the latter approach is a well studied field of
mathematics known as approximation theory, rooted a few centuries
ago in polynomial interpolation of given couples of points, and
developed in statistics through spline
regression~\citep{wahba90splines} and basis
expansions~\citep[\S 5]{hastie01}.

\paragraph{empirical risk minimization:} statistical learning theory starts its course when a probabilistic
knowledge about the generation of the points
$(x,y)$ is assumed, and the reader may refer to~\citep{CucSma02}
for a valuable review. We skip its rigorous exposition, and favour
intuitive arguments next. A sound guess for the learning rule $f$
would be a function with a low empirical risk,
$$R_c^{\emp}\,(\,f\,)\,\defeq\,\frac{1}{n}\sum_{i=1}^n
c\,(f(x_i),y_i),$$ quantified by a cost function
$c:\Ycal\times\Ycal\rightarrow\RR^+$ that penalizes wrong
predictions and which is nil on the diagonal. Minimizing directly
$R_c^{\emp}$ given training sets $X$ and $Y$ is however unlikely
to give interesting functions for $f$. If the function class
$\Fcal$ from which $f$ is selected is large, the problem becomes
ill-posed in the sense that many solutions to the minimization
exist, of which few will prove useful in practice. On the
contrary, if the function class is too restricted, there will be
no good minimizer of the empirical risk that may serve in
practice. To take that tradeoff into account, and rather than
constraining $\Fcal$, assume that $J:\Fcal\rightarrow \RR$ is a
function that quantifies the roughness of a function which is used
to penalize the empirical risk,
\begin{equation}\label{eq:regrisk}
R_c^{\lambda}(f)\defeq \frac{1}{n}\sum_{i=1}^n
c\,(f(x_i),y_i)+\lambda J(f).
\end{equation}
Here $\lambda>0$ balances the tradeoff between two desired
properties for the function $f$, that is a good fit for the data
at hand and a smoothness as measured by $J$. This formulation is
used in most regression and classification settings to select a
good function $f$ as the minimizer of
\begin{equation}\label{eq:program}
\hat{f}=\underset{f\in\Fcal}{\argmin} R_c^{\lambda}.
\end{equation}
\paragraph{kernel classifiers and regressors:} we recover through the formulation of Equation~\eqref{eq:regrisk} a large variety of methods, notably when the penalization is directly related to the norm of
the function in a rkHs:
\begin{itemize}
\item When $\Xcal$ is Euclidian and $\Ycal=\RR$, $\Fcal=\Xcal^*$,
the dual of $\Xcal$ and $c(f(x),y)=(y-f(x))^2$, minimizing
$R_c^\lambda$ is known as least-square regression when
$\lambda=0$; ridge regression~\citep{hoerl1962application} when $\lambda>0$ and $J$ is the
Euclidian 2-norm; the lasso~\citep{tibshirani1996regression} when $\lambda>0$ and $J$ is the
1-norm. \item When $\Xcal=[0,1]$, $\Ycal=\RR$, $\Fcal$ is the
space of $m$-times differentiable functions on $[0,1]$ and
$J=\int_{[0,1]}\left(f^{(m)}(t)\right)^2dt$, we obtain regression
by natural splines of order $m$. This setting actually corresponds
to the usage of thin-base splines which can also be regarded as a
rkHs type method~\citep{wahba90splines}, see~\citep[Table
3]{girosi95regularization} for other examples.

\item When $\Xcal$ is an arbitrary set endowed with a kernel $k$
and $\Ycal=\{-1,1\}$, $\Fcal=\Hcalk$, $J=\|\cdot\|_{\Hcalk}$ and
the hinge loss $c(f(x),y)=(1-yf(x))^+$ is used, we obtain the
support vector machine~\citep{mach:Cortes+Vapnik:1995}. Using the cost function
$c(f(x),y)=\ln(1+e^{-yf(x)})$, yields an extension of logistic regression known as
kernel logistic regression~\citep{zhu_hastie_nips}.

\item When $\Xcal$ is an arbitrary set endowed with a kernel $k$
and $\Ycal=\RR$, $\Fcal=\Hcalk$, $J=\|\cdot\|_{\Hcalk}$ and
$c(f(x),y)=(|y-f(x)|-\varepsilon)^+$, the
$\varepsilon$-insensitive loss function, the solution to this program is known as support
vector regression~\citep{drucker1997support}.
\end{itemize}

Note that by virtue of the representer theorem, recalled above as Theorem~\ref{theo:representer}, that whenever $\Fcal$ is set to be a rkHs $\Hcal$, the mathematical program of Equation~\eqref{eq:program} reaches its minima in the subspace $\Hcal_n$ spanned by the kernel functionals evaluated on the sample points, that is
$$
\hat{f}\in \spa{k(x_i,\cdot)},
$$
hence the function $f$ in Equation~\eqref{eq:regrisk} can be explicitly replaced by a finite expansion \begin{equation}\label{eq:exphcaln}f=\sum_{i=1}^n a_i k(x_i,\cdot),\end{equation} and the corresponding set of feasible solutions $f\in\Hcal$ by $f\in\Hcal_n$ and more simply $a\in\RR^n$ using Equation~\eqref{eq:exphcaln}. The reader may consult~\citep{steinwart2008support} for an exhaustive treatment.
 
\paragraph{kernel graph inference:} we quote another example of a supervized rkHs method. In
the context of supervised graph
inference,~\citet{vert05supervised} consider a set of connected
points $\{x_i\}_{1\leq i\leq n}$ whose connections are summarized
in the combinatorial Laplacian matrix $L$ of their graph, that is
for $i\neq j$, $L_{i,j}=-1$ if $i$ and $j$ are connected and $0$
otherwise, and $L_{i,i}=-\sum_{j\neq i}L_{i,j}$. The authors look
for a sequence of functions $\{f_i\}_{1\leq i\leq d}$ of a rkHs
$\Hcalk$ to map the original points in $\RR^d$, and hope to
recover the structure of the original graph through this
representation. Namely, the projection is optimized such that the
points, once projected in $\RR^d$, will have graph interactions in
that metric (that is by linking all nearest neighbours up to some
distance threshold) that will be consistent with the original
interactions. This leads to successive minimizations that may
recall those performed in kernel-PCA, although different in
nature through the addition of a regularization term proportional to $\lambda$:
$$
f_j=\underset{f\in\Hcalk,f\bot\{f_1,\cdots,f_{j-1}\}}{\argm}
\frac{f_X^T L f_X + \lambda \|f\|_{\Hcalk}}{f_X^T f_X}.
$$
where the vector $f_X$ is defined as
$$f_X\defeq(f(x_1),\cdots,f(x_n))^T.$$ 
The term $f_X^T L f_X$ above can be interpreted as a cost function with respect to the
observable graph $L$, which penalizes functions $f$ that are
for which the values of $f(x_i)$ and $f(x_j)$ are very different for two connected nodes.

\paragraph{kernel discriminant analysis:} we recall briefly the ideas behind the Fisher linear discriminant~\cite{fisherDA} for classification. Given a sample $X=(x_1,\cdots,x_n)$ of points in $\RR^d$ and assume that to each point $x_i$ corresponds a binary variable $y_i\in\{0,1\}$ which is equal to $0$ if $x_i$ belong to a first class and $1$ when $x_i$ belongs to a second class. Fisher discriminant analysis (LDA) assumes that the conditional distributions $p_0(X)=p(X|Y=0)$ and $p_1(X)=p(X|Y=1)$ are both normal densities. If the mean and variances $\mu_0,\mu_1$ and $\Sigma_0$ and $\Sigma_1$ respectively of $p_0$ and $p_1$ were known, the Bayes optimal rule would be to classify any observation $x$ according to the value of its probability ratio $\frac{p_1(x)}{p_0(x)}$ and predict it is in class 0 whenever that ratio is below a certain threshold
$$
(x- \mu_0)^T \Sigma_{0}^{-1} ( x- \mu_0)\ +\ln|\Sigma_{0}|\ -\ (x- \mu_1)^T \Sigma_{1}^{-1} ( x- \mu_1) -\ln|\Sigma_{1}| < \beta. 
$$
If the two classes are homoscedastic, that is $\Sigma_0=\Sigma_1=\Sigma$, then the decision can be simplified to testing whether whether $\omega^Tx < c$ where $\omega$ is defined as $\omega = \Sigma^{-1}(\mu_1-\mu_0)$. This later case is known in the literature as Linear Discriminant Analysis (LDA). When the latter assumption is not valid, Fisher proposed to find a vector $\omega$ that separates the two classes by optimizing the ratio 
$$
r(\omega)= \frac{\left(\omega^T\mu_0 - \omega^T\mu_1\right)^2}{\omega^T\Sigma_0\omega + \omega^T\Sigma_1\omega}= \frac{\left(\omega^T(\mu_0 - \mu_1)\right)^2}{\omega^T\Sigma_0\omega + \omega^T\Sigma_1\omega}
$$
The ratio $r$ is a Rayleigh quotient whose maximum is the only nonzero eigenvalue of the generalized eigenvalue problem $\left(\mu_0 - \mu_1)(\mu_0 - \mu_1)^T,\Sigma_0+\Sigma_1\right)$ which corresponds to the eigenvector
$$
\omega=\left(\Sigma_0+\Sigma_1\right)^{-1}(\mu_0 - \mu_1).
$$
In practice all quantities $\Sigma_i$ and $\mu_i$ are replaced by empirical estimators. As shown in~\cite{mika1999fisher}, the criterion $r$ can be conveniently cast as a quadratic problem in an arbitrary rkHs $\Hcal$ corresponding to a set $\Xcal$. In this new setting, a sample $X=(x_1,\cdots,x_n)$ of $n$ points in $\Xcal$ is paired with a set of labels $(y_1,\cdots,y_n)$. Instead of looking for a vector $\omega$, namely a linear function, kernel discriminant analysis looks for a function $f\in\Hcal_n$ such that 
$$
r(f)= \frac{\left(f(\mu_0) - f(\mu_1)\right)^2}{\var_0 f + \var_1 f},
$$ 
Let us write $n_0$ and $n_1$ for the numbers of elements of $X$ of class $0$ and $1$ respectively, where $n_0+n_1=n$. For functions $f\in\Hcal_n$, namely functions which can be written as $f(\cdot)=\sum_{i=1}^n a_i k(x_i,\cdot)$, we have that
$$
r(f)=  \frac{\left(a^Tm_0 - a^Tm_1\right)^2}{a^TS_0a + a^TS_1a}
$$
where writing 
$$
\begin{aligned}
K &=[k(x_i,x_j)]_{1\leq i,j\leq n},\\ 
K_0 &=[k(x_i,x_j)]_{1\leq i,j\leq n, y_j=0},\\ 
K_1 &=[k(x_i,x_j)]_{1\leq i,j\leq n, y_j=1},
\end{aligned}
$$ 
allows us to express means and variances in of $X$ evaluated in functions of $\Hcal_n$ as 
$$\begin{aligned}
m_0&=K_0\ones_{n_0}, \\
m_1&=K_1\ones_{n_1},\\
S_0&=K_0(I-\frac{1}{n_0}\ones_{n_0mn_0})K_0^T,\\
S_1&=K_1(I-\frac{1}{n_1}\ones_{n_1,n_1})K_1^T.\end{aligned}$$
Following the approach used above for linear functionals in Euclidian spaces, the vector $a$ of weights could be recovered as the (only) nonzero eigenvalue of the $n$ dimensional generalized eigenvalue problem $\left((m_0+m_1)(m_0+m_1)^T,(S_0+S_1)\right)$. However, as the reader may easily check, the matrix $S_0+S_1$ is not-invertible in the general case. Adding a regularization term $\lambda I_n$ is a hack that makes the problem computationally tractable. It can also be motivated from a regularization point of view. Indeed, since we are looking to maximize the ratio, this modification is equivalent to adding the rkHs norm of $f_\alpha$ to its denominator and hence favor functions with low norm. Although this explanation is not as motivated as the empirical regularization scheme discussed in Section~\ref{sec:supervized}, it is the one provided in the original work of~\citet{mika1999fisher}. Note in particular that the represent theorem does not apply in this setting and hence looking for function in $\Hcal_n$ is itself an arbitrary choice. The kernel discriminant is thus the function $f$ such that 
$$
f(\cdot)=\sum_{i=1}^na_i k(x_i,\cdot),\;\;\;\; a=\left(S_0+S_1+\lambda I_n \right)^{-1}(m_0+m_1).
$$

\subsection{Density Estimation and Novelty Detection}
A density estimator is an estimator based on a data sample of points drawn independently and identically distributed according to an unobservable underlying probability density function. The level sets of the estimator are the sets of points in $\Xcal$ for which the density of the estimator has values below or above a given threshold. Estimating level sets rather than a density estimator taken within a set of candidate densities is the nonparametric direction taken by the one-class support vector machine presented below.

\paragraph{one-class SVM:} taking advantage of the support vector machine formulation to minimize the penalized empirical risk of Equation~\eqref{eq:regrisk},~\citet{scholkopf99ocsvm} proposed the reformulation
$$
R_c^{\lambda}(f)\defeq \frac{1}{n}\sum_{i=1}^n c\,(f(x_i))+ \nu \|f\|_\Hcal.
$$
where the labels of all points are set to 1 to estimate a function $f$ that is positive on its support and that takes smaller values on areas of lower densities. $c$ can be any convex function differentiable at $0$ and such that $c'(0)<0$. In particular,~\citet{scholkopf99ocsvm} solves the following mathematical program
$$
\BA{ll}
\mbox{minimize} & \nu \|f\|_\Hcal + \sum_{i=1}^n (\xi_i - \rho) \\
\mbox{subject to}& f\in\Hcal_n\\
& f(x_i)\leq \rho-\xi_i,\xi_i\geq 0
\EA
$$

\paragraph{novelty detection and kernel-PCA:}
Novelty detection refers to the task of detecting patterns in a given data set that do not conform to an established normal behavior~\citep{chandola09anomaly}. novelty detection can be implemented in practice by using the level sets of a density estimator. 
A new observation is intuitively labelled as abnormal if it lies within a region of low density of the estimator granted this new observation has been drawn from the same distribution. Another approach to novelty detection is given by a spectral analysis implemented through the study of the principal components of a data sample. This approach can be naturally generalized to a ``kernelized algorithm''.

Principal component analysis can be used as a novelty detection tool for multivariate data~\cite[\S10.1]{jolliffe:PCA} assuming the underlying data can be reasonably approximated by a Gaussian distribution. Given a sample $X=(x_1,\cdots,x_n)$ of $n$ points drawn i.d.d from the distribution of interest in $\RR^d$, the $p$ first eigenvectors of PCA are defined as the $p$ first orthonormal eigenvectors $e_1,\cdots,e_p$ with corresponding eigenvalues $\lambda_1,\cdots,\lambda_p$ of the sample variance matrix $\Sigma_n=\frac{1}{n-1}\sum (x_i-m)(x_i-m)^T$ where $m=\frac{1}{n}\sum_{i=1}^n x_i$ is the sample mean. An observation $y$ is labelled as abnormal whenever its projection in the space spanned by the $p$ eigenvectors is markedly outside the ellipsoid defined by the semi-axes $(e_i,\lambda_i)$, namely when
$$\sum_{i=1}^p \frac{e_i^T(y-m)}{\lambda_i^2}\geq \beta,$$ or alternatively when the contribution of the first $p$ eigenvectors to the total norm of $y$ is low compared to the weight taken by the other directions of lower variance,
$$\frac{\|y\|^2-\sum_{i=1}^p \frac{e_i^T(y-m)}{\lambda_i^2}}{\|y\|^2}\geq \alpha.$$
This idea has been extended in the case of kernel-PCA in~\citep{hoffmann2007kpn} by using a kernel $k$ on $\Xcal$. In that case the linear functionals $e_i^T\cdot$ are replaced by the evaluations of eigenfunctions $f_i(\cdot)$ introduced in Equation~\eqref{eq:kpcadef} and the norm of $y$ itself is taken in the corresponding rkHs $\Hcal$ and can be recovered as $k(y,y)$.

\newpage\section{Kernel Selection and Kernel Mixture}\label{chap:interactions}
An important issue that arises when using kernel machines in practice is to select an adequate kernel. In most practical cases choices are abundant if not infinite. We review three different families of techniques designed to cope with this situation. The following section is set in the usual classification setting where the dataset of interest is composed of $n$ pairs of points and labels, i.e. $\{(x_i,y_i)\}_{i=1..n}$ where each $x_i\in \Xcal$ and $y_i\in \{-1,1\}$.

\subsection{Parameter Selection}\label{sec:paramsel}
When the kernel can be parameterized by a few variables, a brute force approach that examines the cross-validation error over a grid of acceptable parameters is a reasonable method which often yields satisfactory results. This approach is non-tractable when the number of parameters reaches but a few values. ~\citep{chapelle:2002,bousquet2003complexity,flich2004feature} and more recently~\citep{keerthi2007efficient} have proposed different schemes to tune the parameters of a Gaussian kernel on $\RR^d$. The authors usually assume a setting where the weights $\sigma_i$ assigned to each feature of two vectors $x$ and $y$ in $\RR^d$ need to be tuned, that is consider kernels of the form
$$
k(x,y)=\exp\left(-\sum_{i=1}^d \frac{(x_i-y_i)^2}{\sigma_i^2}\right).
$$
Finding an adequate parameter choice $(\sigma_1,\cdots,\sigma_d)$ implies defining first a criterion to discriminate good from bad choices for such parameters. ~\citet{chapelle:2002} consider the leave-one-out error of the regularized empirical risk formulation of Equation~\eqref{eq:regrisk}. Given a kernel $k_\sigma$ parameterized by a set of parameters $(\sigma_1,\cdots,\sigma_d)$, the leave-one-out error is the sum
$$
\Ecal_{\text{LOO}}(\sigma)=\frac{1}{n}\sum \ones(f_{-i}(x_i)\ne y_i),
$$ 
where we use a generic regularized empirical risk minimizer estimated on all points of the sample but one:
$$f_{-i}=\argmin_{f\in\Hcal} \frac{1}{n}\sum_{j=1, j\ne i }^n
c\,(f(x_j),y_j)+\lambda J(f).$$
\citet{evgeniou2004leave} show that the leave-one-out error is a good way to quantify the performance of a class of classifiers. If $\Ecal_{\text{LOO}}$ was a tractable and analytical function of $\sigma$, it would thus seem reasonable to select $\sigma$ as a minimizer of $\Ecal_{\text{LOO}}$. This is not the case unfortunately. The authors of both~\citep{chapelle:2002,bousquet2003complexity} propose to consider instead upperbounds on $\Ecal_{\text{LOO}}$ which are tractable and design algorithms to minimize such upperbounds through gradient descent methods.~\citet{keerthi2007efficient} generalize this approach by considering other proxies of the performance of a kernel on a given problem.

\subsection{Multiple Kernel Learning}\label{subsec:addmix}
Rather than looking for a single kernel in a large set of candidates, a research trend initiated by~\citet{lanckriet2004} proposes to consider instead combinations of candidate kernels. As recalled in Section~\ref{sec:distndpd} positive definite kernels can be combined multiplicatively (under point-wise multiplication) and linearly (through positive linear combinations).
Since the pioneering work of~\cite{lanckriet2004}, which relied on expensive semi-definite programming to compute optimal linear combinations of kernels, the shift of study has progressively evolved towards computationally efficient alternatives to define useful additive mixtures as in~\citep{bach2004skm,SonRaeSchSch06,rakoto2007}. A theoretical foundation for this line of research can be found in~\cite{micchelli2006learning}. We follow the exposition used in~\citep{rakoto2007}. Recall, as exposed in Section~\ref{sec:supervized}, that given a kernel $k$, kernel classifiers or regressors yield decision functions of the form
\begin{equation}\label{eq:finalfunction}
f(x) = \sum_{i=1}^{n} \alpha_i^\star y_i k(x, xi) + b^\star,
\end{equation}
where both the family $(\alpha_i^\star)$ and $b^\star$ stand for optimized parameters. When not one, but a family of $m$ kernels $k_1,\ldots,k_m$ kernels can be combined in a convex manner to yield a composite kernel $k=\sum_{l=1}^m d_l k_l$ with $\sum d_l=1$, the task consisting in learning both the coefficients $\alpha_i,b$ and the weights $d_l$ in a single optimization problem is
known as the multiple kernel learning (MKL) problem~\citep{bach2004skm}. Writing $\Hcal$ for the rkhs corresponding to kernel $k$, The penalized SVM-type optimization framework for the estimation of a function $f$ in $\Hcal$ is
$$
\BA{ll}
\mbox{minimize} & \|f\|^2_{\Hcal} + C \sum{\xi_i}\\
\mbox{subject to} & f\in \Hcal, \;\;b\in\RR,\\
& \forall i, y_i (f(x_i)+b) \geq 1 - \xi_i, \\
& \xi_i \geq 0.
\EA
$$
When the kernel $k$ is a mixture of $m$ kernels, the authors propose the following optimization scheme,
$$
\BA{ll}
\mbox{minimize} & \sum_l \frac{1}{d_l}\|f_l\|^2_{\Hcal_l} + C \sum{\xi_i}\\
\text{subject to} &f_l\in \Hcal_l, \;\;b\in\RR, d_l, \xi_i,\\
& \forall i\leq n,\; y_i (\sum_l f_l(x_i)+b) \geq 1 - \xi_i,\\ &\sum_l d_l =1 ;\;\; \xi_i,d_l \geq 0 .
\EA
$$
In the formulation above, each value $d_l$ controls the importance given to squared norm of $f_l$ in
the objective function. A bigger $d_l$ favors functions whose component in $\Hcal_l$ may have a larger norm. If the weight $d_l$ goes to zero, the corresponding function $f_l$ can only be zero as shown by the authors, which is indeed equivalent to not taking into account kernel $k_l$ in the sum $\sum_{k} d_l k_l$. The solution to the problem above can actually be decomposed into a two-step procedure, namely by minimizing an objective function $J(d)$ defined on the weights $d=(d_l)$ and which is itself computed through a SVM optimization, namely:
\begin{equation}\label{eq:mkltop}
\BA{ll}
\mbox{minimize}& J(d),\\
\mbox{subject to} & \sum_{l}d_l=1; \;d_l\geq 0,\\& J(d) =\BA[t]{ll}\mbox{minimize}  &\sum_l \frac{1}{d_l}\|f_l\|^2_{\Hcal_l} + C \sum{\xi_i}\\
\mbox{subject to }& f_l\in \Hcal_l, b\in\RR, \xi_i\\ &\forall i, y_i (\sum_l f_l(x_i)+b) \geq 1 - \xi_i ; \xi_i\geq 0.\EA
\EA
\end{equation}
The authors iterate between the computation of the objective function $J$, itself a SVM optimization, and the optimization of $J$, carried out using projected Gradient methods. Each iteration of this loop involves the computation of the Gradient's directions $\frac{\partial J}{\partial d_l}$ which the authors show are simple functions of the weights $\alpha_i^\star$ retrieved during the SVM-computation conducted to compute $J$, namely
$$
\frac{\partial J}{\partial d_l} = -\frac{1}{2} \sum_{i,j=1}^n \alpha_i^\star \alpha_j^\star y_i y_j k_l(x_i,x_j)
$$
The algorithm boils down to the following loop.
\begin{itemize}
\item initialize all weights $d_l$ to $1/m$,
\item Loop :\begin{itemize}
\item compute an SVM-solution to the problem with fixed weights $d$. This gives $J$, as well as its associated Gradient directions $\frac{\partial J}{\partial d_l}$.
\item Optimize $J$ with respect to $d$, that is replace the current weights family $d$ by $d + \gamma D$ where $D$ is the vector of descent direction computed from the Gradient (reducing it and projecting it) such that the new $d$ satisfies the simplex constraints, and $\gamma$ is an optimal step size determined by line search.
\item Check for optimality conditions initially set, and if reached get out of the loop. \end{itemize}
\end{itemize}

By the end of the convergence, both the weights $\alpha$ and $b$ that arise from the last computation of $J$, that is the SVM-computation step, and the weights $d$ obtained in the end provide the parameters needed to define $f$ in Equation~\eqref{eq:finalfunction}. An additional property of the algorithm is that it tends to produce sparse patterns for $d$, which can be be helpful to interpret which kernels are the most useful for the given task.
\subsection{Families of Kernels Labeled on a Graph}
The approach taken in the latter section assumes that all kernels are independently selected. The optimization of Equation~\eqref{eq:mkltop} is carried out on the linear subspace formed by all linear combinations of these $m$ kernels. Rather than
treating all kernels uniformly and mixing them linearly,~\citep{cuturi2006kernels} consider a setting with two particular features.

\begin{figure}[htbp]
\includegraphics[width=12cm]{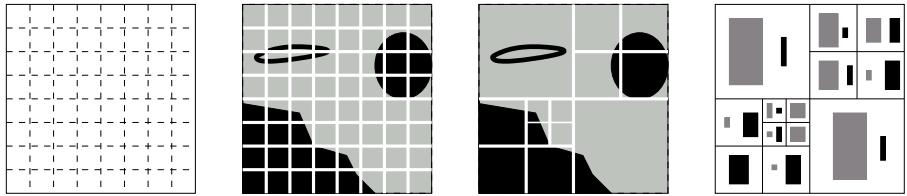}
\caption{The leftmost image represents the final nodes of the hierarchy considered, that is the $4^3$ squares present in the grid. The ancestors of such nodes correspond iteratively to the $16$ larger square
obtained when grouping $4$ small windows, then to the image divided
into $4$ equal parts and finally the whole image. The hierarchy has thus a depth of 3. Any
set of nodes taken in the hierarchy can in turn be used to compare two images under the light of those local color histograms displayed in the right most image, which reduces in the case of
two-color images to binary histograms as illustrated in the
right-most image.}\label{fig:multiresgrid}
\end{figure}

First, an a-priori knowledge on the structure on the kernels themselves can be used, namely a hierarchical structure under the form of a tree. Kernels are indexed by labels on a directed acyclic graph (DAG) $\{\alpha\in \Tcal\}$, and each kernel $k_\alpha$ is related to its siblings, that is  $k_\beta,\beta\in s(\alpha)$ where $s(\alpha)$ stands for the sons of a node $\alpha$. An example of such an approach can be seen in Figure~\ref{fig:multiresgrid} where the hierarchy is a dyadic partition of the surface of an image.

Second, the hierarchy can be used not only to combine kernels additively but also multiplicatively. More precisely, the authors define the space of candidate kernels as the space $\Scal$ of all complete subtrees $t$ of $\Tcal$ starting with the same root. Such a tree $t$ is uniquely characterized by its set of final nodes $f(t)$, and the kernel associated to such a subtree is the product of the kernels associated to each final node, that is
    $$
    k_t = \prod_{\alpha\in f(t)}k_\alpha.
    $$
Note that the number of potential subtrees grows super-exponentially, hence yielding a number of candidate kernels far superior to the total number of node kernels.

Grounded on these two assumptions,~\citet{cuturi2006kernels} use a prior weight on subtree kernels to propose a fast computation of a kernel $k$ as
$$
k = \sum_{t\in\Scal} d_t k_t.
$$
The weight $d_t$ penalizes the complexity of a given subtree $t$ by considering its number of nodes. In practice the weights $d_t$ are defined with an analogy to branching process priors for trees~\citep{jagers1975branching}.~\citet{francis2008} proposed a similar setting to optimize directly the weights $d_t$ using a variation of the Multiple Kernel Learning framework. The originality of the approach is to take advantage of the hierarchy between kernels to adaptively explore subtrees which may fit better the task at stake.

\newpage\section{Kernel Cookbook}\label{chap:cookbook}
We review in this section practical guidelines that can apply to the selection of a kernel given a dataset of structured objects.
\subsection{Advanced Kernels on Vectors}
Vectors of $\RR^d$ can be directly approached by considering the linear dot-product as a kernel, which
amounts to performing an alternative penalized regression and optimizing
it in the dual space as is described in~\cite{chapelletraining}. 

Beyond the use of the linear kernel, the array of positive definite kernels defined on vectors if not on scalars is very large, and include functions of all possible shapes as illustrated in~\citep[Exercise 2.12, p.79]{berg84harmonic}. Although some specific kernels have been used for their precise invariance properties~\cite{fleurettriangular},
most practitioners limit themselves to the use of Gaussian and
polynomial kernels. Once a family of kernel has
been selected, the topic of choosing adequate parameters for this kernel
is itself one of the biggest challenges when using kernel methods on
vectorial data, as hinted in~\citep[Section
12.3.4]{hastie01}. For polynomial kernels searches are usually limited
to the offset and exponent parameters. In the case of Gaussian kernels,
usually favored by practitioners, the more general use of Mahalanobis
distances instead of the simple Euclidian distance, that is kernels of
the form
$$
k_\Sigma(x,y)=e^{-\frac{1}{2}(x-y)^T\Sigma(x-y)}
$$
where $\Sigma$ is a $d\times d$ symmetric positive definite matrix, has
also been investigated to fit better data at hand and to insist on the
possible correlations or importance of the described features. The
simplest type of matrices $\Sigma$ which can be used is one with a
diagonal structure, and pre-whitening the data might be considered as
such an approach. More advanced tuning strategies have been covered in Section~\ref{sec:paramsel}.
\subsection{Kernels on Graphs}
Labeled graphs are widely used in computer science to model
data linked by discrete dependencies, such as social networks, molecular pathways or patches in images. Designing kernels for graphs is usually done with this wide applicability in mind.

A graph $G$ is described by a finite set of vertices $\Vcal$ and a hence
finite set of edges $E= \Vcal \times \Vcal$. Graphs are sometimes labelled. In that case there exists a function of $E$ to the set of labels $\Lcal$, or alternatively $\Vcal$ to $\Lcal$ that assigns a label to a node or an edge. 

Given an arbitrary subgraph $f$ and a graph of interest $G$, the feature $f(G)$ measuring how many
subgraphs of $G$ have the same structure as graph $f$ is a useful elementary feature. The original paper by~\citet{KasTsuIno03} presented in Section~\ref{chap:smk} uses for the set of subgraphs $f$ simple random walks and counts their co-occurrences to provide a kernel, an approach that had also been studied in the case of trees~\citep{vert-tree}. The work has found extensions in~\citep{mahe2005} to take better into account similarity between not only the graph structure but also the labels that populate it. More advanced sets of features, which rely on algebraic descriptions of graphs have been recently considered in~\citep{DBLP:conf/icml/KondorSB09}. We refer the reader to the exhaustive review of~\citet{vishwanathan2008graph}.

\subsection{Kernels on Images}
Technically speaking, an image can be seen as a long 3-dimensional vector of RGB intensities. It is however unlikely that treating images as vectors and applying Gaussian kernels on them will yield any interesting result. In that sense, the definition of kernels for images are build on higher-level properties of images and images contents, such as the invariance to slight translations in both color intensities and patterns positions in the image. These properties can be translated into the following kernels.

\begin{figure}
\begin{center}
\includegraphics[height=8.5cm,width=11.5cm]{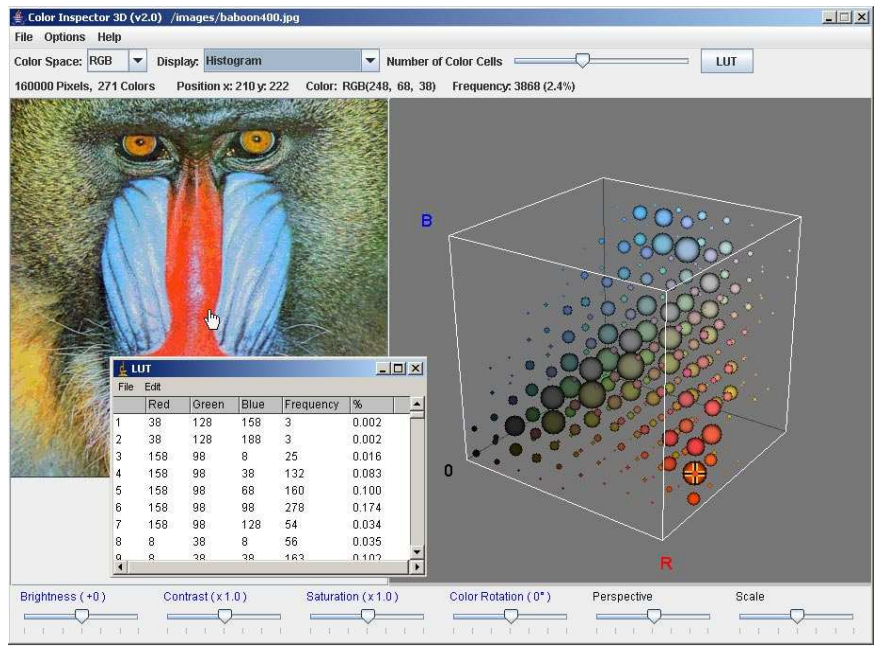}
\caption{A complex image such as the monkey above can be summarized through color histograms, represented, as above, as a 3D histogram of red/green/blue intensities. Although this representation inquires considerable loss of information, it is often used for image retrieval.}\label{fig:monkey}
\end{center}
\end{figure}

\paragraph{color histograms:}
numerous approaches have stemmed from the use of color histograms to build kernel on images, starting with the seminal experiments carried out by~\citet{chapelle99SVMs}. By representing an image $I$ by an arbitrary color histogram $\theta_I\in\Sigma_d$, where $d$ stands for the color depth (typically 256, 4096 or 16 million), the authors follow by designing a kernel on two images using kernels on multinomials such as those presented in Section~\ref{sec:vectors}, typically
$$
k(I_1,I_2)= e^{-\lambda \|\theta_{I_1}-\theta_{I_2}\|}.
$$
Note that this approach assumes a \emph{total invariance} under pixel translation, which is usually a drastic loss of information on the structure and the content of the image itself, as illustrated in Figure~\ref{fig:monkey}.

Further developments have tried to cope with this this limitation. Rather than considering a single histogram for each image ~\citet{conf/iccv/GraumanD05} and \citet{cuturi2006kernels} divide the image into local patches and compare the resulting families of local histograms. These approaches provide substantial improvements at a low computational cost.

\paragraph{taking shapes taken into account:}
note first that by counting elementary shapes in each image, rather than colours, the techniques described in the paragraph above can be applied to define kernels that focus on shape similarities. However, when larger shapes are the main criterion to discriminate two images, histogram representations have obvious limitations.~\citet{icpr_HaasdonkK02} propose a kernel which exponentiates the opposite of the tangent distance between two images to quantify their similarity. Since the computation of the tangent distance requires the optimization of a criterion, the framework is related to other attempts at designing a kernel from of distance, e.g \citep{watkins00dynamic,nips02-AA20,VerSaiAku04,cuturi07permanents,cuturi07kernel}. The tangent distance~\citep{Simard98transformationinvariance} is a distance computed between two shapes $x$ and $y$ to assess how different they are to each other by finding an optimal series of elementary transformations (rotations, translations) that produces $y$ when starting from $x$. Since the distance is not negative-definite the tangent distance does not yield directly a positive definite kernel, but might be used in practice with most kernel machines after an adequate correction.

\paragraph{shapes seen as graphs:}
taking such limitations into account but still willing to incorporate a discrimination based on shapes, \citet{DBLP:conf/cvpr/HarchaouiB07} have exploited existing graph kernels to adapt them to images. Images can indeed be seen as large graphs of interconnected color dots.~\citet{DBLP:conf/cvpr/HarchaouiB07} propose to segment first the images through standard techniques (in the quoted paper the authors use the watershed transform technique) into large areas of homogeneous colors, and then treat the resulting interconnections between colored areas as smaller graphs labeled with those simplified colors. The two graphs are subsequently compared using standard graph kernels, notably a variation proposed by the authors. When the number of active points in the images is low,~\citet{conf/icml/Bach08} focus on a specific category of graph kernels tailored for point clouds taking values in 2D or 3D spaces.

\subsection{Kernels on Variable-Length Sequential Data}
Variable-length sequence data-types are ubiquitous in most machine learning applications. They include the observation sampled from a discrete-time processes, texts as well as long strings such as protein and DNA codes. One of the challenges of designing kernels on such objects is that such kernels should be able to compare sequences of different lengths, as would be the case when comparing two speech segments with different sampling frequencies or overall recorded time, two texts, or two protein with different total number of amino acids.

 \paragraph{kernels on texts:}
most kernels used in practice on texts stem from the use of the popular bag-of-words (BoW) representations, that is sparse word count vectors taken against very large dictionaries. The monograph~\citep{joachims:2002a} shows how the variations of the BoW can be used in conjunction with simple kernels such as the ones presented in Section~\ref{sec:vectors}. From a methodological point of view, much of the approach relies rather on choosing efficient BoW representations and on the contrary usually boil down to the use of simple kernels.

\paragraph{histograms of transitions:}
when tokens are discreet and few, the easiest approach is arguably to map them as histograms of shorter substrings, also known as $n$-grams, and compare those histograms directly. This approach was initially proposed by~\citep{leslie02spectrum} with subsequent refinements to either incorporate more knowledge about the tokens transitions~\citep{cuturi05context,leslie03mismatch} or improve computational speed~\citep{Teo06fastand}.

\paragraph{higher level transition modeling with HMM's:} rather than using simple $n$-gram counts descriptors, ~\citet{jaakola00discriminative} use more elaborate statistical models to define kernels between strings which can be modelled as HMM. The interested reader may refer to Section~\ref{chap:smk} for a review of the Fisher kernel to see how the HMM model is used to build a feature vector to compare strings directly.

 \paragraph{edit distances:}
a different class of kernels can be build using transformations on the sequences themselves, in a form that echoes with the Tangent distance kernel presented in an earlier section. Intuitively, if by successive and minor changes one can map a sequence $x$ to another sequence $y$, then the overall cost (which remains to be defined) needed to go from $x$ to $y$ can be seen as a good indicator of how related they are to each other. As with the tangent distance reviewed above, ~\citet{shimodaira02dynamic} take into account the optimal route from $x$ to $y$, whose total cost is known depending on the application field as the edit distance, the Smith-Waterman score, Dynamic-Time-Warping or Levenshtein distance, to define a kernel which is not necessarily positive-definite but which performs reasonably well on a speech discrimation task. ~\citet{VerSaiAku04} argue that by taking a weighted average of the costs associated to all possible transformations mapping $x$ to $y$, one can obtain a kernel who is positive definite and which usually performs better on the set of proteins they consider in their study. Up to a few subtleties, a similar approach is presented in~\citep{cuturi07kernel} which shows good performance on speech data.

\newpage
\bibliographystyle{apa}

\end{document}